\PassOptionsToPackage{table,dvipsnames}{xcolor}
\documentclass{article} 
\usepackage[preprint]{colm2025_conference}

\usepackage{amsmath}

\usepackage[utf8]{inputenc} 
\usepackage[T1]{fontenc}    
\usepackage{hyperref}       
\usepackage{url}            
\usepackage{booktabs}       
\usepackage{amsfonts}       
\usepackage{nicefrac}       
\usepackage{microtype}      
\usepackage{xcolor}         

\usepackage{tabularx}
\usepackage{multirow}
\usepackage{makecell}
\usepackage{graphicx} 
\usepackage{wrapfig}
\usepackage{enumitem}
\usepackage{tcolorbox}
\usepackage{listings} 
\tcbuselibrary{listings}

\newtcblisting{datasetbox}[2][]{
  title={\textbf{#2}},
  colback=gray!5,
  colframe=gray!50,
  fonttitle=\bfseries,
  listing only,
  listing options={breaklines=true, basicstyle=\ttfamily\small},
  #1
}

\newtcblisting{promptbox}[2][]{
  title={\textbf{#2}},
  colback=blue!5,
  colframe=blue!50!black,
  fonttitle=\bfseries,
  listing only,
  listing options={breaklines=true, basicstyle=\ttfamily\small},
  #1
}

\newtcblisting{listbox}[2][]{
  title={\textbf{#2}},
  colback=cyan!5,
  colframe=cyan!60!black,
  fonttitle=\bfseries,
  listing only,
  listing options={breaklines=true, basicstyle=\ttfamily\small},
  #1
}

\title{Reinforced Reasoning for Embodied Planning}

\author{%
  Di Wu \\
  Tongji University \\
  \texttt{diwu7012@gmail.com} \\
  \And
  Jiaxin Fan\textsuperscript{*} \\
  Tongji University \\
  \texttt{2253538@tongji.edu.cn} \\
  \And
  Junzhe Zang\textsuperscript{*} \\
  Tongji University \\
  \texttt{2250724@tongji.edu.cn} \\
  \And
  Guanbo Wang \\
  Tsinghua University \\
  \texttt{wanggb23@mails.tsinghua.edu.cn} \\
  \And
  Wei Yin \\
  Bank of Communications \\
  \texttt{yinw\_8@bankcomm.com} \\
  \And
  Wenhao Li\textsuperscript{\dag} \\
  Tongji University \\
  \texttt{whli@tongji.edu.cn} \\
  \And
  Bo Jin\textsuperscript{\dag} \\
  Tongji University \\
  \texttt{bjin@tongji.edu.cn} \\
}

\begin{document}

\maketitle

\begingroup
\renewcommand\thefootnote{}\footnotetext{
\textsuperscript{*}Equal contribution. \quad
\textsuperscript{\dag}Corresponding authors.
}
\endgroup

\begin{abstract}
  Embodied planning requires agents to make coherent multi-step decisions based on dynamic visual observations and natural language goals. While recent vision-language models (VLMs) excel at static perception tasks, they struggle with the temporal reasoning, spatial understanding, and commonsense grounding needed for planning in interactive environments. In this work, we introduce a reinforcement fine-tuning framework that brings R1-style reasoning enhancement into embodied planning. We first distill a high-quality dataset from a powerful closed-source model and perform supervised fine-tuning (SFT) to equip the model with structured decision-making priors. We then design a rule-based reward function tailored to multi-step action quality and optimize the policy via Generalized Reinforced Preference Optimization (GRPO). Our approach is evaluated on Embench, a recent benchmark for interactive embodied tasks, covering both in-domain and out-of-domain scenarios. Experimental results show that our method significantly outperforms models of similar or larger scale, including GPT-4o-mini and 70B+ open-source baselines, and exhibits strong generalization to unseen environments. This work highlights the potential of reinforcement-driven reasoning to advance long-horizon planning in embodied AI. We released all code and data at \href{https://github.com/mail-taii/Reinforced-Reasoning-for-Embodied-Planning}{https://github.com/mail-taii/Reinforced-Reasoning-for-Embodied-Planning}.
\end{abstract}

\section{Introduction}\label{introduction}
\begin{wrapfigure}{r}{0.6\textwidth}  
  \vspace{-10pt}                     
  \centering
  \includegraphics[width=0.58\textwidth]{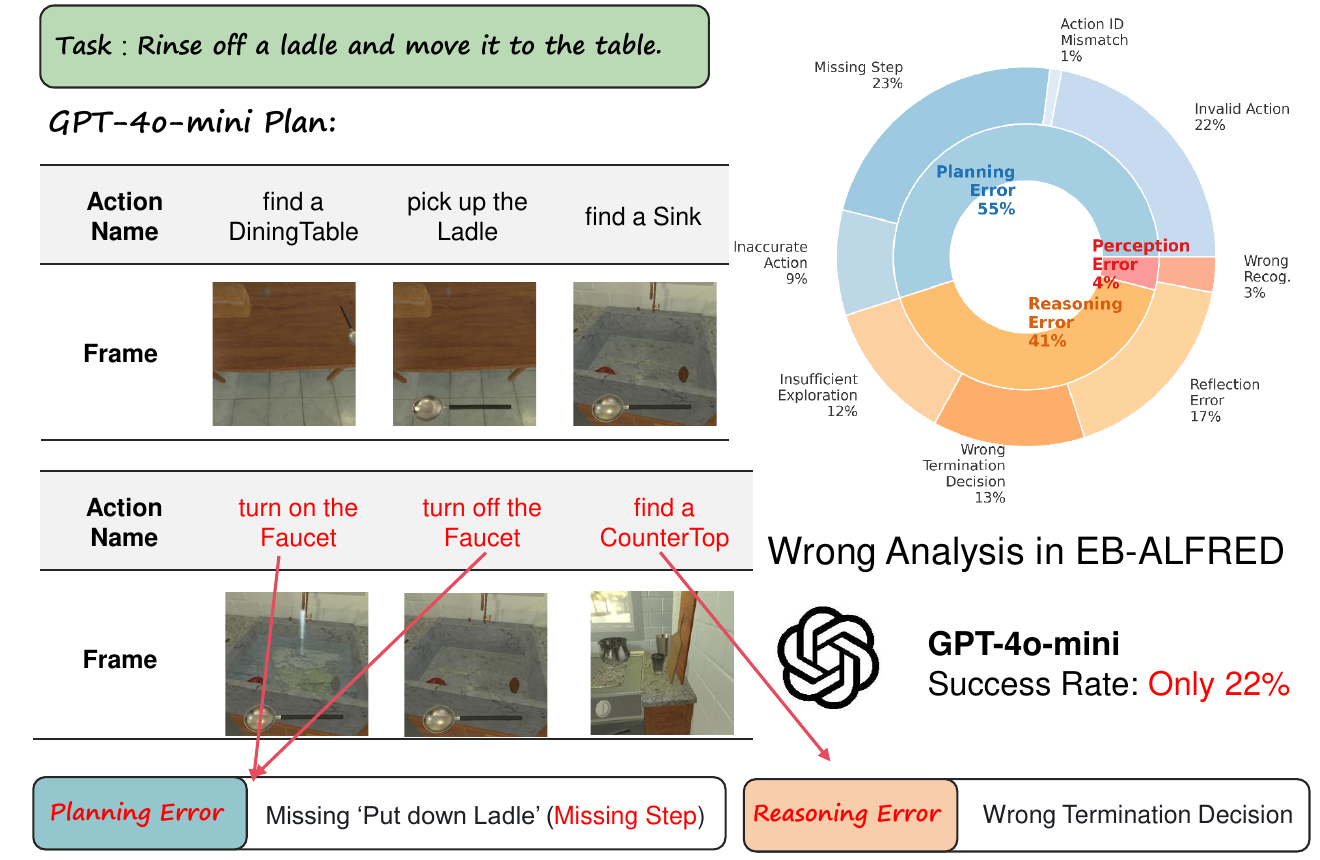}
  \vspace{-6pt}                      
  \caption{Failure case and error breakdown of GPT-4o-mini in the EB-ALFRED environment. 
\textbf{Left:} A representative task failure. 
\textbf{Right:} Distribution of failure types across EB-ALFRED tasks.}
  \label{fig:wrap_pipeline}
\end{wrapfigure}

Embodied planning serves as a cornerstone in hierarchical embodied AI systems\cite{shi2025hi,zhang2024hirt}, where intelligent agents must not only perceive their environment but also reason and act within it to accomplish complex, real-world tasks\cite{duan2022survey}. Unlike low-level controllers that govern precise trajectory execution\cite{ECoT,kimopenvla}, high-level planning is responsible for formulating coherent action sequences that translate complex instructions into manageable sub-tasks\cite{tapa}. While conventional language-based reasoning is confined to static, text-driven contexts\cite{lightman2023let,ye2025limo,shao2024deepseekmath}, embodied planning operates within dynamic, interactive environments that demand sequential decision-making across multiple steps. Despite recent advancements in Vision-Language Models (VLMs) have demonstrated impressive capabilities in static understanding tasks\cite{zhang2024vision}, they exhibit substantial limitations when applied to multi-step interactive embodied planning. Empirical analyses in Figure\ref{fig:wrap_pipeline} reveal that even state-of-the-art VLMs, which excel in image captioning or visual question answering, struggle to maintain coherent and efficient decision sequences in dynamic environments\cite{yang2025embodiedbench}. These shortcomings highlight a critical gap: effective planning in real-world embodied contexts imposes far greater demands on spatial reasoning, temporal consistency, and commonsense understanding than current VLM architectures can satisfy.

To address reasoning deficiencies, recent research has explored enhancing large models' cognitive abilities through dedicated reasoning frameworks\cite{plaat2024reasoning}. Notably, approaches such as DeepSeek-R1\cite{guo2025deepseek} have pioneered reinforcement-driven paradigms that explicitly strengthen a model’s reasoning capacity via reward-guided optimization, and have achieved promising results in math and code problems. Extensions of this paradigm into multimodal contexts have begun to emerge\cite{wang2025multimodal}, tackling tasks such as visual mathematics and diagram-based reasoning\cite{r1-vl,vlm-r1,meng2025mm,visualRFT}. However, applying such reasoning-enhancement techniques to embodied planning remains highly challenging and underexplored due to the fundamental differences between embodied tasks and conventional reasoning benchmarks: (1) Embodied planning requires spatial perception and physical commonsense\cite{ma2024survey}, whereas tasks like math or code focus purely on symbolic reasoning without grounding in dynamic environments; (2) The transition from static, single-turn QA to interactive, multi-turn decision-making\cite{wang2025ragen} introduces continuous feedback loops—unlike static tasks, embodied agents must adaptively reason as each action reshapes their environment; (3) Embodied planning lacks unique ground-truth trajectories, in contrast to deterministic domains, as multiple valid solutions exist for a single goal, complicating reward design and supervision. 

In this work, we bridge the gap by proposing a reinforcement fine-tuning framework that brings R1-style reasoning enhancement into embodied planning, enabling models to make more coherent and context-aware decisions in dynamic, interactive environments. We propose a rule-based reward function that specifically designed for multi-step decision, and optimize the model using Generalized Reinforced Preference Optimization (GRPO) \cite{shao2024deepseekmath} to encourage long-horizon, goal-directed reasoning. Prior to reinforcement learning, we distill response patterns from a large closed-source model to construct a high-quality training corpus and perform supervised fine-tuning (SFT)\cite{SFT}, equipping the model with rich commonsense priors and structured reasoning habits as a foundation for downstream optimization. Recognizing the discrepancy between simplistic text-based simulations and the complexities of real-world physics, we conduct evaluations within Embench\cite{yang2025embodiedbench}, an interactive embodied benchmark that faithfully captures environmental dynamics and agent-environment feedback loops. Experimental results demonstrate that our method significantly improves planning performance, yielding more efficient and context-aware action sequences. Moreover, our reinforcement-driven fine-tuning exhibits strong generalization across unseen tasks and environments, underscoring its potential for practical deployment in real-world embodied AI applications.

In summary, our contributions are as follows:
\begin{itemize}[leftmargin=*]
  \item We are the first to apply reinforcement fine-tuning to optimize a vision-language model for embodied planning, significantly improving the model’s ability to perform coherent multi-step reasoning and decision-making in dynamic environments.
  \item We propose a comprehensive training pipeline that integrates supervised fine-tuning (SFT) with reinforcement fine-tuning (RFT), alongside carefully constructed datasets, a reward function tailored for multi-step decision-making, and supporting mechanisms such as online data filtering, leading to consistent and robust performance improvements.
  \item We conduct extensive evaluation on Embench, an interactive benchmark for embodied AI, showing that our model not only outperforms comparable-scale models but also surpasses GPT-4o-mini and open-source models with more than 70B parameters. It further demonstrates strong generalization to unseen domains, validating the generality of reinforcement-based adaptation.
\end{itemize}

\section{Related Work} 
\subsection{Embodied Task Planning}
Embodied task planning focuses on decomposing high-level natural language instructions into executable sequences of sub-tasks, enabling agents to perform complex behaviors in interactive environments. With the emergence of large language and vision-language models\cite{xi2025rise,xu2024survey}, researchers have explored using pretrained LLMs or VLMs to generate plans from textual and visual observations, typically relying on carefully crafted prompts\cite{shin_socratic_2024,rana_sayplan_nodate,hu_look_2023,kim_context-aware_2024,singh_progprompt_2022,fu2024can} or auxiliary tools\cite{rana_sayplan_nodate,saycan,silver_generalized_2024} to provide necessary planning cues. While simple and data-efficient, such methods often struggle with spatial grounding and temporal coherence in visually rich environments. Advanced methods have tried to fine-tune LLMs or VLMs to improve planning performance. Several works have employed supervised fine-tuning pipelines\cite{tapa,chen_robogpt_2024,ji_robobrain_2025}, while others adopt preference optimization methods\cite{wang_world_2025,song2024trial} such as Direct Preference Optimization (DPO)\cite{dpo} to better align model behavior with expert planning preferences.

Despite these advances, most existing methods operate in static or offline settings, where plans are generated without actual interaction with the environment. In this work, we address this limitation by evaluating our model in interactive environments\cite{yang2025embodiedbench} bridging the gap between static planning capabilities and dynamic embodied execution.

\subsection{Vision-Language Model Reasoning}
Reasoning in vision-language models (VLMs) involves drawing inferences from both textual and visual inputs, often requiring spatial, temporal, or causal understanding\cite{wang2025multimodal,wang2024exploring}. A common approach is Chain-of-Thought (CoT) prompting\cite{wei2022chain}, where the model generates intermediate steps to clarify its reasoning. In multimodal settings, Multimodal Chain-of-Thought (MCoT) extends this idea by integrating visual inputs like images and videos into the reasoning process\cite{zhang2023multimodal,mondal2024kam,mitra2024compositional}.

More recently, R1-style reinforcement learning\cite{guo2025deepseek,shao2024deepseekmath} has emerged as an effective framework for enhancing reasoning capabilities. These methods optimize reasoning quality through reward-guided learning, enabling models to self-correct and generate more detailed reasoning processes. Originally developed for text-based reasoning, R1 approaches have since been extended to multimodal domains, including image-based QA\cite{visualRFT,vlm-r1,r1-vl}, visual math problems\cite{meng2025mm,vision-r1,team2025kimi}, and video reasoning\cite{li2025videochat}. In the context of embodied AI, some early studies\cite{embodiedR,EmbodiedReasoner,reasonrft} have applied R1-based training to question answering tasks, however, they primarily focus on short-horizon QA tasks. In contrast, our work is the first to adopt R1-style reinforcement fine-tuning for long-horizon embodied planning, aiming to improve structured decision-making across multiple interactive steps.

\section{Methodology}\label{method}
\begin{figure}[!t]
\centering
    \includegraphics[width=\linewidth]{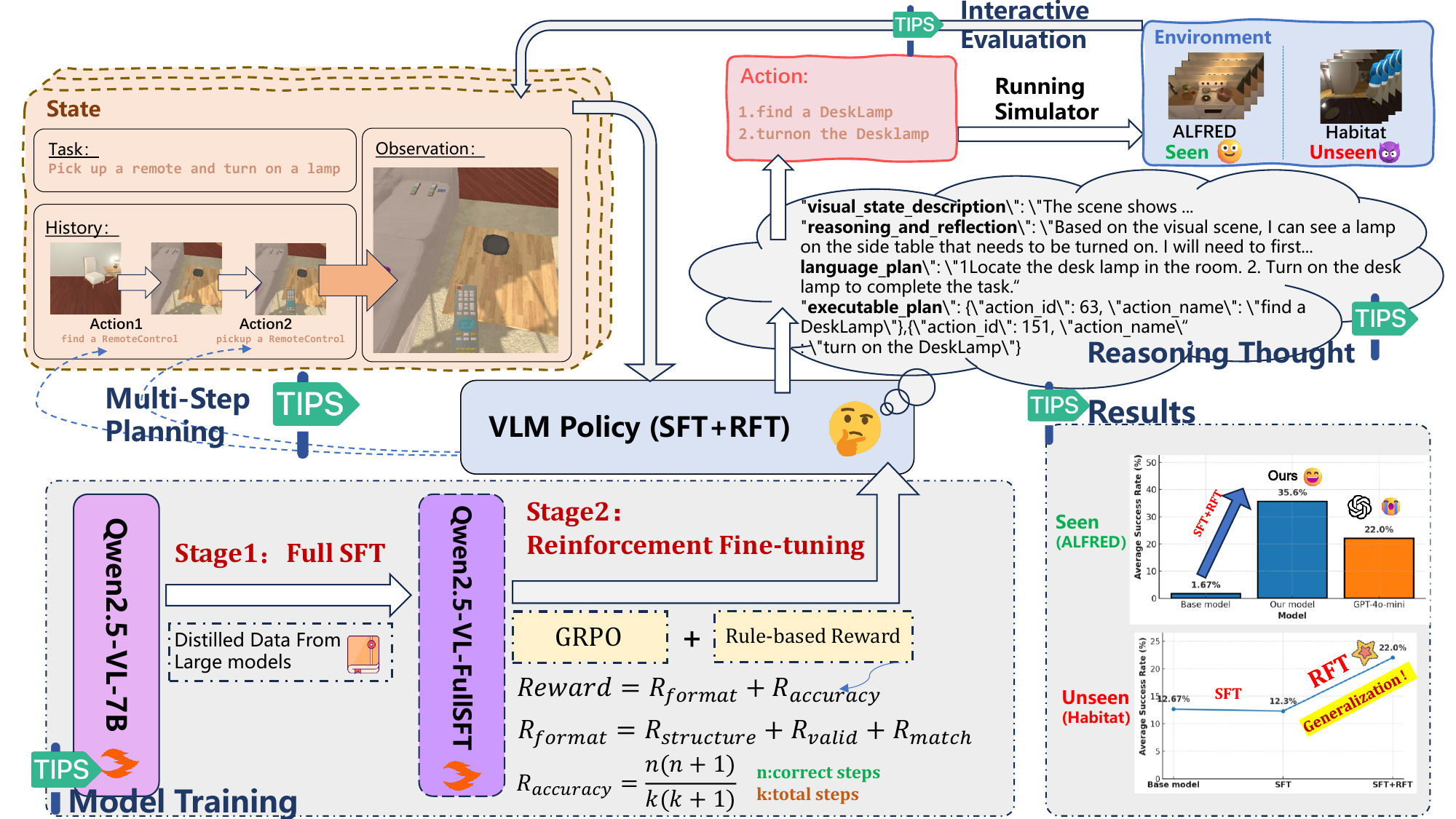}
    \caption{Overview of our proposed framework. We adopt a two-stage training paradigm consisting of supervised fine-tuning (SFT) followed by reinforcement fine-tuning (RFT) to enhance multi-step planning capabilities of the vision-language model. The final model is evaluated on \textbf{Embench}, an interactive embodied benchmark, where it achieves strong performance across both \textbf{seen }and \textbf{unseen} environments.}
    \label{fig:overall_large_pic}
\end{figure}
\subsection{Problem Definition}

We formulate embodied task planning as a partially observable decision-making process, where the agent interacts with an environment through sequential actions based on visual observations. At each time step $t$, the agent receives an observation $o_t \in \mathcal{O}$ and executes an action $a_t \in \mathcal{A}$, forming a history

\begin{equation}
h_t = \{o_0, a_0, o_1, ..., o_t\}.
\end{equation}

Given a task instruction $g \in \mathcal{G}$ described by a natural language command $L$, the task is associated with a set of binary goal-checking conditions $\mathcal{C}(g) = \{c_1, ..., c_k\}$ that must all be satisfied for the task to be considered successful. The agent generates a trajectory 

\begin{equation}
e = (g, o_0, a_0, o_1, ..., o_n, a_n),
\end{equation}

and the reward is defined as
\begin{equation}
r(e) = \mathbb{I} \left[ \bigwedge_{c \in \mathcal{C}(g)} c = \text{True} \right],
\end{equation}

where $\mathbb{I}[\cdot]$ is the indicator function.

We parameterize the policy $\pi_\theta$ using a vision-language model (VLM), which outputs an action distribution conditioned on the observation $o_t$, history $h_t$, instruction $L$, and a fixed prompt template $P$:

\begin{equation}
a_{t+1} \sim \pi_\theta(\cdot \mid o_t, h_t, L, P).
\end{equation}

Our objective is to optimize $\theta$ such that the expected task success rate of sampled trajectories increases:

\begin{equation}
\max_\theta \ \mathbb{E}_{e \sim \pi_\theta} \left[ r(e) \right].
\end{equation}

We adopt a two-stage training paradigm: supervised fine-tuning (SFT) to align $\pi_\theta$ with high-quality trajectories, followed by reinforcement fine-tuning (RFT) to further improve performance under interactive evaluation.

\subsection{Preparing for Reinforcement: Distilled Supervised Fine-tuning}

Embodied planning requires strong spatial perception and commonsense reasoning abilities. However, small and open-source vision-language models (VLMs) often fall short in these aspects compared to proprietary large-scale models. To bridge this capability gap and provide a solid initialization for subsequent reinforcement fine-tuning, we first adopt supervised fine-tuning (SFT)\cite{SFT} on a high-quality dataset obtained via large model distillation.

\paragraph{Distillation from Large Models.}  
Unlike tasks with well-defined ground-truth labels, embodied planning allows for diverse valid trajectories to accomplish the same goal. Collecting human-annotated demonstrations for each trajectory is labor-intensive and lacks scalability. Therefore, we opt for a distillation-based approach: we prompt a proprietary model, Gemini-2.0-flash\cite{gemini2.0}, to solve embodied planning tasks and record its outputs to construct our SFT dataset.

Specifically, for each task goal $g \in \mathcal{G}$ and environment observation history $h_t$, we construct a prompt $p = \texttt{Prompt}(g, h_t)$, and collect Gemini's response $\hat{a}_{t+1}$. The dataset is represented as a collection of tuples:

\begin{equation}
\mathcal{D}_{\text{SFT}} = \left\{ (p_i, \hat{a}_i) \right\}_{i=1}^{N},
\end{equation}

where $p_i$ is the textual input prompt and $\hat{a}_i$ is the response generated by Gemini, which contains both the planning trajectory and associated reasoning process. In total, we collect over 4000 training samples and use them to supervise the open-source Qwen2.5-VL\cite{qwen2.5} model.

\paragraph{Supervised Fine-tuning.}  
Given the distilled dataset $\mathcal{D}_{\text{SFT}}$, we optimize the model parameters $\theta$ of the VLM policy $\pi_\theta$ via maximum likelihood estimation:

\begin{equation}
\mathcal{L}_{\text{SFT}}(\theta) = - \mathbb{E}_{(p, \hat{a}) \sim \mathcal{D}_{\text{SFT}}} \left[ \log \pi_\theta(\hat{a} \mid p) \right].
\end{equation}

We explore both full-parameter fine-tuning and parameter-efficient LoRA-based\cite{hu2022lora} fine-tuning strategies. Empirically, we observe that full fine-tuning yields slightly better performance.

Overall, the SFT stage enables the model to inherit the task-decomposition patterns, commonsense priors, and spatial grounding demonstrated by the larger model. Details on training configuration and dataset content are provided in Appendix.

\subsection{Reinforcing Reasoning for Embodied Planning}


While SFT improves task-specific performance, it often lacks the reasoning generalization needed for unseen scenarios. Recent work such as DeepSeek-R1\cite{guo2025deepseek} shows that reinforcement learning (RL) with rule-based rewards can effectively enhance reasoning by optimizing for quality over imitation, improving both task success and generalization—especially important in embodied contexts.

Building on this, we propose a reinforcement fine-tuning framework for long-horizon embodied planning. Unlike prior RL methods limited to short-horizon QA, we extend to interactive multi-step tasks. We construct a dataset from the ALFRED benchmark\cite{alfred}, design a rule-based reward for planning evaluation, and optimize the VLM using the GRPO algorithm\cite{shao2024deepseekmath}, with an online filtering strategy to improve training stability.

\paragraph{Dataset Construction.}

The visual reinforcement fine-tuning dataset consists of samples formatted to support reward-based optimization. Each sample is represented as a triplet $(L, o, \hat{a})$, where $L$ denotes the textual input instruction, $o$ is the image input, and $\hat{a}$ is the ground-truth answer used for reward computation. For multi-step planning, we decompose each reference trajectory $e = (g, o_0, a_0, o_1, a_1, ..., o_k, a_k)$ of length $k$ into $k$ training samples. At each step $n \in [1, k]$, we construct $L_n$ by embedding the task goal $g$ along with the previous action history $a_{0:n-1}$. The corresponding observation $o_n$ is taken from the $n$-th step, and the target $\hat{a}_{n:} = \{a_n, a_{n+1}, ..., a_k\}$  consists of the remaining actions from step $n$.

We build this dataset on top of the ALFRED benchmark\cite{alfred}, which provides complete execution traces of agents performing household tasks in a simulated environment. Applying the above trajectory decomposition strategy, we collect a total of 43,898 training samples.


\paragraph{Reward Function.}

Inspired by prior work in reinforcement fine-tuning\cite{meng2025mm,guo2025deepseek}, we design a composite reward function that integrates both format correctness and action accuracy. The goal is to guide the model toward producing structured, valid, and effective multi-step plans. We denote the total reward as:

\begin{equation}
R(\text{response}, \text{answer}) = R_{\text{format}}(\text{response}) + R_{\text{accuracy}}(\text{response}, \text{answer}),
\end{equation}

\textbf{(1) Format Reward.}  
Unlike prior works that rely on generic templates such as \texttt{<think>} and \texttt{<action>}, we tailor the reward to suit the structured output required for embodied multi-step planning. The response is expected to contain a JSON object with specific keys: \texttt{reasoning\_and\_reflection}, \texttt{visual\_state\_description}, \texttt{language\_plan}, and \texttt{executable\_plan}. This output structure is inspired by the Embench\cite{yang2025embodiedbench} prompting format, where the model is encouraged to first observe the image, then reflect and reason, and finally produce a coherent multi-step action plan. The format reward is computed as:

\begin{equation}
R_{\text{format}} = R_{\text{structure}} + R_{\text{valid}} + R_{\text{match}},
\end{equation}

where:
\begin{itemize}
  \item $R_{\text{structure}} = 0.125$ if all required top-level fields exist, otherwise $0$.
  \item $R_{\text{valid}} = 0.125 \times \frac{\text{\# valid steps}}{\text{\# total steps}}$, where a step is valid if it contains an integer \texttt{action\_id} and a string \texttt{action\_name}.
  \item $R_{\text{match}} = 0.25 \times \frac{\text{\# correctly matched actions}}{\text{\# total steps}}$, where a match is counted only if the pair \texttt{(action\_id, action\_name)} corresponds to a valid and correct entry in the predefined action mapping, ensuring consistency and preventing hallucinated actions.

\end{itemize}

\textbf{(2) Accuracy Reward.}  
To assess execution correctness, we compare the predicted action sequence $\hat{a} = \{a_1, ..., a_k\}$ to the reference (gold) action sequence $a^* = \{a_1^*, ..., a_k^*\}$. The comparison is performed step-by-step in a prefix-matching manner: starting from the first step, each predicted action must exactly match the corresponding ground-truth action. Once a mismatch is encountered, the comparison stops. Let $n$ denote the number of consecutively matched steps, i.e., the prefix length such that $a_i = a_i^*$ for all $i \in [1, n]$.

The accuracy reward is defined as:
\begin{equation}
R_{\text{accuracy}} = R(n; k),
\end{equation}

where $R(n; k)$ denotes the multi-step reward allocation curve described in the next section. Additionally, for single-step tasks ($k = 1$), if the model generates more than one step, we apply a penalty of $-0.25$ to discourage redundant actions.

\textbf{(3) Multi-step Reward Allocation Curve.}  
To reflect long-horizon planning quality, we define a progressive reward allocation curve that assigns higher reward to longer correct prefixes. Given a reference sequence of length $k$ and a matched prefix of length $n \leq k$, we compute the reward using triangular normalization:
\begin{equation}
R(n; k) = \frac{n(n+1)}{k(k+1)}.
\end{equation}

This function grows quadratically with $n$ and is normalized to the range $[0, 1]$, assigning proportionally more reward for longer correct planning. It encourages models not only to predict the correct final outcome, but also to maintain consistency and correctness throughout the entire action sequence.

\paragraph{Optimization Method.}

We adopt Group Relative Policy Optimization (GRPO) \cite{shao2024deepseekmath} to optimize the VLM policy under reward-based supervision.Given a prompt $x$ , the policy model $\pi_\theta$ generates a set of $G$ sampled responses $\{y_1, y_2, \dots, y_G\} \sim \pi_\theta(\cdot \mid x)$. Each response $y_i$ is scored by the reward function $r_i = R(y_i)$ that reflects its quality in terms of format and planning accuracy. GRPO computes the relative advantage $A_i$ of each response as its normalized deviation from the group mean:
\begin{equation}
A_i = \frac{r_i - \operatorname{mean}(\{r_1, \dots, r_G\})}{\operatorname{std}(\{r_1, \dots, r_G\})},
\end{equation}

The training objective encourages the model to increase the likelihood of high-quality responses under the current policy while maintaining stability with respect to a reference policy $\pi_{\text{ref}}$. The GRPO loss is defined as:

\begin{equation}
\mathcal{J}(\theta) = \mathbb{E}_{x \sim \mathcal{D}} \ \mathbb{E}_{\{y_i\} \sim \pi_\theta} \left[ \frac{1}{G} \sum_{i=1}^{G} \left( \operatorname{clip} \left( \frac{\pi_\theta(y_i \mid x)}{\pi_{\text{old}}(y_i \mid x)},\ 1-\epsilon,\ 1+\epsilon \right) \cdot A_i - \beta \cdot \mathcal{D}_{\mathrm{KL}}(\pi_\theta \| \pi_{\text{ref}}) \right) \right],
\end{equation}

where $\pi_{\text{old}}$ is the policy used for sampling, $\epsilon$ controls the clipping range, and $\beta$ is the weight for the KL penalty between the current and reference policies.

This optimization objective enables stable and lightweight training by leveraging relative preferences within sampled groups, without the need for absolute reward values or additional critic networks.

\paragraph{Data Filtering Strategy.}

In the early stages of reinforcement fine-tuning, we observe that many sampled responses from the model yield extremely low reward values, leading to weak or unstable gradient signals. To address this issue and ensure stable policy updates, we incorporate an online data filtering strategy followed by PRIME\cite{cui2025process} and MM-Eureka\cite{meng2025mm}. The core idea is to discard samples that are either too poor or too perfect to maintain informative and diverse training batches throughout optimization.Concretely, we apply a filtering criterion based on response-level accuracy. For each input prompt $x$, we generate a group of $G$ responses $\{y^{(i)}\}_{i=1}^{G}$ from the current policy $\pi_\theta$, and compute their individual rewards $r^{(i)} = R(x, y^{(i)})$. We then define the accuracy of a prompt group as:

\begin{equation}
C_x = \left| \{ y^{(i)} \mid r^{(i)} = 1 \} \right|,
\end{equation}

which counts how many responses in the group achieve full reward. We retain a prompt group for training only if its accuracy falls within a predefined range: $\epsilon_{\text{acc}}^{\text{lower}} \leq C_x \leq \epsilon_{\text{acc}}^{\text{upper}}$, ensuring that the group contains a balanced mix of good and poor responses.

Accepted samples are buffered into a memory set $\mathcal{B}$ of size $N_B$. Once the buffer is filled, we perform $K_2$ steps of GRPO optimization on the collected data, after which the buffer is cleared and the process repeats.This filtering mechanism significantly improves learning stability by eliminating gradient degeneracy and encourages the policy to learn from relatively informative contrastive examples.

\section{Experiments} \label{exp}
{%
  \setlength{\textfloatsep}{-4pt}
  {%
  \setlength{\extrarowheight}{2pt}  
  \setlength{\tabcolsep}{0pt}       

  \begin{table}[htbp]
    \scriptsize                   
    \centering

    \begin{tabular*}{\textwidth}{@{\extracolsep{\fill}}
        >{\centering\arraybackslash}p{2.5cm} 
        >{\centering\arraybackslash}p{0.6cm}
        *{7}{>{\centering\arraybackslash}p{0.7cm}}
    }
      \toprule
      \textbf{Model} 
        & \textbf{Params.} 
        & \multicolumn{7}{c}{\textbf{EB-ALFRED (Seen)}} \\
      \cmidrule(lr){3-9}
      & 
        & \textbf{Avg} 
        & \textbf{Base} 
        & \textbf{Common} 
        & \textbf{Complex} 
        & \textbf{Visual} 
        & \textbf{Spatial} 
        & \textbf{Long} \\
      \midrule

      \rowcolor{orange!15}
      \multicolumn{9}{c}{\textbf{Closed-Source MLLMs}} \\
      Claude-3.5-Sonnet          & —  &  66.0  & 70 & 62 & 72 & 62 & 60 & 70 \\
      Gemini-2.0-flash           & —  &  51.3  & 58 & 58 & 50 & 46 & 42 & 54 \\
      GPT-4o                     & —  &  54.6  & 62 & 52 & 68 & 44 & 48 & 54 \\
      GPT-4o-mini                & —  &  22.0  & 32 & 24 & 32 & 20 & 24 & 0 \\

      \rowcolor{orange!15}
      \multicolumn{9}{c}{\textbf{Open-Source General MLLMs}} \\
      Llama-3.2-11B              & 11B  &   13.3    &   22  &   8  &   16  &   22  &   6  &   6  \\
      Qwen2.5-VL-7B              & 7B &  1.7  & 4 & 2 & 2 & 2 & 0 & 0 \\
      InternVL2.5-8B             & 8B &  3.0  & 2 & 0 & 12 & 0 & 4 & 0 \\

      \rowcolor{orange!15}
      \multicolumn{9}{c}{\textbf{Open-Source Reasoning MLLMs}} \\
      R1-VL-7B           
                                  & 7B &   2    &  2  &  2  &  6  &  0  &  0  &  2  \\
      \makecell[c]{MM-Eureka-Qwen-7B}
                                  & 7B & 2.67  & 6 & 4 & 4 & 2 & 0 & 0 \\

      \rowcolor{orange!15}
      \multicolumn{9}{c}{\textbf{Open-Source Embodied MLLMs}} \\
      RoboBrain                  & 7B  & 0.33  & 2 & 0 & 0 & 0 & 0 & 0 \\
      Tapa                       & 7B  &   0    &   0  &   0  &   0  &   0  &   0  &   0  \\

      \rowcolor{orange!15}
      \multicolumn{9}{c}{\textbf{Open-Source Embodied + Reasoning MLLMs}} \\
      Ours                       & 7B  &   35.6    &   54  &   42  &   46  &   28  &   38  &   6 \\

      \bottomrule
    \end{tabular*}

    \caption{Performance Comparison of Models on EB-ALFRED (Seen)}
    \label{tab:eb-alfred-seen}
  \end{table}
}%

  \begin{figure}[htb!]
    \begin{minipage}[t]{0.53\textwidth}
      \vspace{0pt}                             
      \raggedright
      \includegraphics[
          width=\linewidth,
          trim=0 0 0 28pt, clip               
      ]{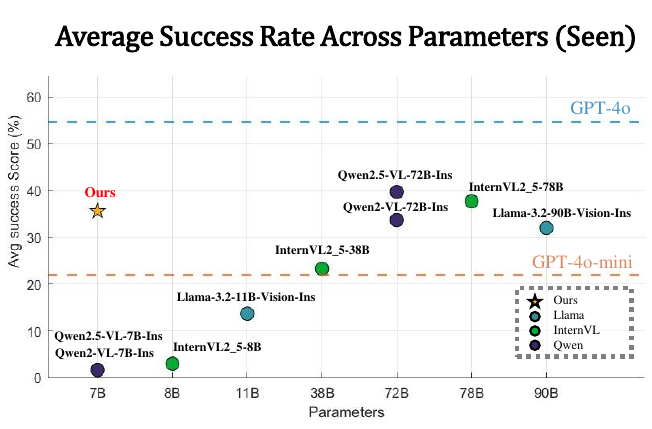}
      \caption{Success rate vs.\ parameters \textbf{(Seen)}}
      \label{fig:alfred_scatter}
    \end{minipage}
    \hspace{2pt}                               
    \begin{minipage}[t]{0.45\textwidth}
      \vspace{0pt}
      \raggedright
      \includegraphics[width=\linewidth]{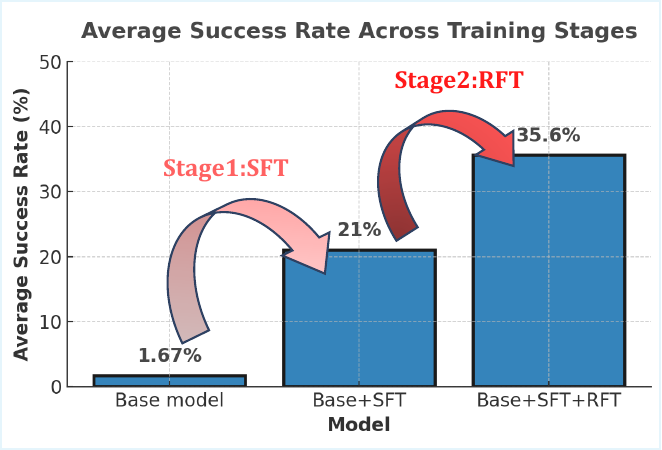}
      \caption{Success rate vs.\ stages \textbf{(Seen)}}
      \label{fig:alfred_bar}
    \end{minipage}
  \end{figure}
}

\subsection{Embodied Multi-step Planning Evaluation in Interactive Environment}

\subsubsection{Benchmark}

Most prior works in embodied planning reduce evaluation to static visual question answering, which fails to capture the interactive and sequential nature of real-world decision-making. To address this gap, we adopt \textbf{Embench}\cite{yang2025embodiedbench}, a benchmark designed for evaluating multimodal agents in dynamic, interactive environments.

Embench provides a unified framework across four embodied settings and supports over 1,100 tasks involving manipulation, navigation, and spatial reasoning. We evaluate on two environments: \textbf{EB-ALFRED}, built on ALFRED\cite{alfred} and AI2-THOR\cite{ai2thor}, and \textbf{EB-Habitat}, based on Habitat 2.0’s rearrangement tasks\cite{habitat}. Tasks are categorized into six subsets: \textit{Base}, \textit{Common Sense}, \textit{Complex Instruction}, \textit{Spatial Awareness}, \textit{Visual Appearance}, and \textit{Long Horizon}, enabling fine-grained capability analysis.

All models generate step-by-step plans from egocentric inputs and execute them in simulation. Since our training data is collected from the ALFRED simulator, EB-Habitat serves as an \textbf{out-of-domain} setting for generalization evaluation. More details are provided in Appendix.

\subsubsection{Baselines}
We compare our method against a range of baselines, including: (1) proprietary models such as Claude-3.5-Sonnet\cite{claude3.5}, Gemini-2.0-flash\cite{gemini2.0}, GPT-4o\cite{gpt-4o}, and GPT-4o-mini\cite{gpt-4o-mini}; (2) open-source general VLMs like LLaMA-3.2-Vision-11B\cite{llama}, Qwen2.5-VL-7B\cite{qwen2.5}, and InternVL2.5-8B\cite{intervl}; (3) reasoning-oriented models such as MM-Eureka\cite{meng2025mm} and R1-VL\cite{r1-vl}; and (4) embodied VLMs including RoboBrain\cite{ji_robobrain_2025} and TAPA\cite{tapa}. For evaluation, we convert visual inputs into text for TAPA due to its lack of vision capabilities. Further details on each baseline are provided in Appendix.

{%
  \setlength{\extrarowheight}{3pt}  
  \setlength{\tabcolsep}{6pt}       

  \begin{table}[htbp]
    \scriptsize                   
    \centering

    \begin{tabular*}{\textwidth}{@{\extracolsep{\fill}}
        >{\centering\arraybackslash}p{4cm} 
        >{\centering\arraybackslash}p{0.6cm}
        *{7}{>{\centering\arraybackslash}p{0.7cm}}
    }
      \toprule
      \textbf{Model} 
        & \textbf{Params.} 
        & \multicolumn{7}{c}{\textbf{EB-Habitat (Unseen)}} \\
      \cmidrule(lr){3-9}
      & 
        & \textbf{Avg} 
        & \textbf{Base} 
        & \textbf{Common} 
        & \textbf{Complex} 
        & \textbf{Visual} 
        & \textbf{Spatial} 
        & \textbf{Long} \\
      \midrule

      \rowcolor{green!12}
      \multicolumn{9}{c}{\textbf{Closed-Source MLLMs}} \\
      Claude-3.5-Sonnet          & —  &  67.7  & 96 & 68 & 74 & 74 & 40 & 54 \\
      Gemini-2.0-flash           & —  &  34.3  & 76 & 30 & 30 & 30 & 26 & 14 \\
      GPT-4o                     & —  &  54.0  & 82 & 34 & 62 & 58 & 32 & 56 \\
      GPT-4o-mini                & —  &  32.3  & 68 & 38 & 28 & 28 & 22 & 10 \\

      \rowcolor{green!12}
      \multicolumn{9}{c}{\textbf{Open-Source General MLLMs}} \\
      Llama-3.2-11B              & 11B  &   23.3    &   62  &   16  &   24  &   14  &   18  &   6  \\      
      Qwen2.5-VL-7B              & 7B &  12.7  & 38 & 4 & 12 & 4 & 12 & 6 \\
      InternVL2.5-8B             & 8B &  17.0  & 48 & 6 & 16 & 10 & 18 & 4 \\

      \rowcolor{green!12}
      \multicolumn{9}{c}{\textbf{Open-Source Reasoning MLLMs}} \\
      R1-VL-7B          
                                  & 7B &   7.3  &  24   &  0   &  4   &  6   &  8   &  2   \\
      \makecell[c]{MM-Eureka-Qwen-7B}
                                  & 7B &  16.3  & 40 & 16 & 14 & 10 & 16 & 2 \\

      \rowcolor{green!12}
      \multicolumn{9}{c}{\textbf{Open-Source Embodied MLLMs}} \\
      RoboBrain                  & 7B  &  15.3  & 38 & 6 & 18 & 8 & 18 & 4 \\
      Tapa                       & 7B  &    0    &  0   &  0   &  0   &  0   &  0   &  0   \\

      \rowcolor{green!12}
      \multicolumn{9}{c}{\textbf{Open-Source Embodied + Reasoning MLLMs}} \\
      Ours                       & 7B  &    20    &  56   &  8   &  18   &  16   &  14   &  8   \\

      \bottomrule
    \end{tabular*}

    \caption{Performance Comparison of Models on EB-Habitat (Unseen)}
    \label{tab:eb-habitat}
  \end{table}
}%

\begin{figure}[htb!]
  \begin{minipage}[t]{0.55\textwidth}
    \raggedright                   
    \includegraphics[width=\linewidth]{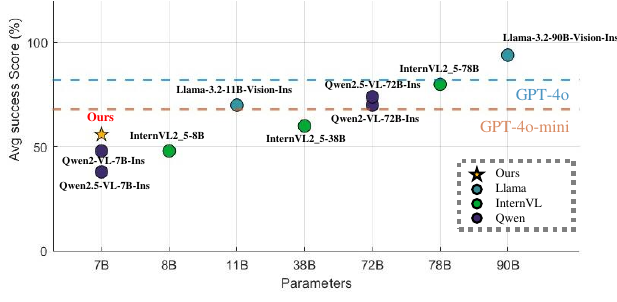}
    \caption{Success rate vs. parameters \textbf{(Unseen)}}
    \label{fig:habitat_scatter}
  \end{minipage}
  \hfill                            
  \begin{minipage}[t]{0.45\textwidth}
    \raggedright
    \includegraphics[width=\linewidth]{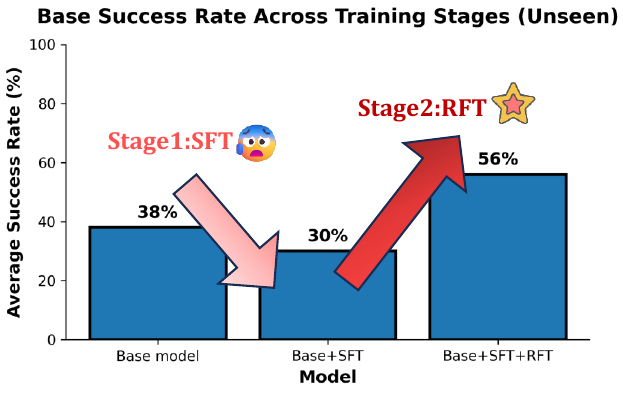}
    \caption{Success rate vs. stages \textbf{(Unseen)}}
    \label{fig:habitat_bar}
  \end{minipage}
\end{figure}

\subsubsection{Implementation Details}

For supervised fine-tuning (SFT), we train our model using the LLaMA-Factory\cite{zheng2024llamafactory} framework on 4 NVIDIA A100 40G GPUs for approximately 8 hours, using Qwen2.5-VL-7B\cite{qwen2.5} as base model. For reinforcement fine-tuning (RFT), we use the OpenRLHF\cite{hu2024openrlhf} framework and perform GRPO optimization on 8 A100 40G GPUs, with one training epoch requiring approximately 40 hours.

For evaluation, all models are deployed via Flask-based inference servers. Each model is evaluated on both EB-ALFRED and EB-Habitat environments, with end-to-end evaluation time per model being around 18 hours. For reasoning-heavy models such as R1-VL and MM-Eureka, the inference latency is significantly higher, resulting in total evaluation time of up to 2–3× longer. 


\subsection{Experiment Results}

\subsubsection{In-Domain Results}

We conduct comprehensive in-domain evaluations on the EB-ALFRED environment. As shown in Table~\ref{tab:eb-alfred-seen}, Figure~\ref{fig:alfred_scatter}, and Figure~\ref{fig:alfred_bar}, our proposed model achieves a task success rate of 35.6\%, significantly outperforming GPT-4o-mini (22.0\%) and much larger models such as Qwen2.5-VL-72B (33.7\%) and LLaMA3.2-90B-Vision-Ins(32.0\%).

Several key observations emerge from the results: (1) Our two-stage training pipeline (SFT + RFT) leads to consistent performance gains in embodied task planning. (2) Existing open-source reasoning models and embodied VLMs perform poorly in Embench. While reasoning models produce verbose intermediate steps, they struggle to execute correct action sequences. Similarly, embodied VLMs lack the generalization ability to transfer to Embench tasks. (3) Long-horizon tasks remain a major challenge. Despite overall improvement on other categories, the performance gain in Long-Horizon tasks is marginal, highlighting the need for future research on planning depth and temporal reasoning.

\subsubsection{Out-of-Domain Results}

To evaluate generalization, we tested our models in the EB-Habitat environment, which differs from ALFRED in terms of scenes, objects, action space, and task types. As shown in Table~\ref{tab:eb-habitat}, our method exhibits strong out-of-domain performance, outperforming all baseline models of similar 7B size, including general-purpose, reasoning-augmented, and embodied VLMs.

We highlight the following findings: (1) Reinforcement fine-tuning leads to substantial improvements even in completely unseen environments, validating the cross-domain robustness of our approach,in contrast, supervised fine-tuning alone offers no benefit in out-of-domain settings. (2) Since our training dataset is constructed primarily from base instructions in ALFRED, the improvement is more pronounced in Base-type tasks within EB-Habitat, while gains in other categories remain limited. This observation suggests the need for more diverse training data to support broader generalization.

\begin{table}[!t]
  \centering
  \begin{minipage}[t]{0.48\linewidth}
    \vspace{0pt}
    \scriptsize
    \setlength{\extrarowheight}{2pt}
    \setlength{\tabcolsep}{4pt}
    \centering
    \begin{tabular*}{\linewidth}{@{\extracolsep{\fill}}lcccc}
      \toprule
      \multirow{2}{*}{\textbf{Variant}} 
        & \multicolumn{2}{c}{\textbf{EB-ALFRED (Seen)}} 
        & \multicolumn{2}{c}{\textbf{EB-Habitat (Unseen)}} \\ 
      \cmidrule(lr){2-3}\cmidrule(l){4-5}
      & \textbf{Avg} & \textbf{Base} & \textbf{Avg} & \textbf{Base} \\
      \midrule
      Base                         & 1.67 & 4  & 12.67 & 38 \\
      SFT only                     & 21   & 34 & 12.3  & 30 \\
      RFT only                     & 9.3  & 18 & 15.6   & 40 \\ 
      RFT\,$\rightarrow$\,SFT      & 29   & 40 & 10.3  & 30 \\
      SFT\,$\rightarrow$\,RFT (ours) & \textbf{35.6} & \textbf{54} & \textbf{20} & \textbf{56} \\
      \bottomrule
    \end{tabular*}
  \end{minipage}
  \hfill
  \begin{minipage}[t]{0.48\linewidth}
    \vspace{0pt}
    \centering
    \includegraphics[width=\linewidth]{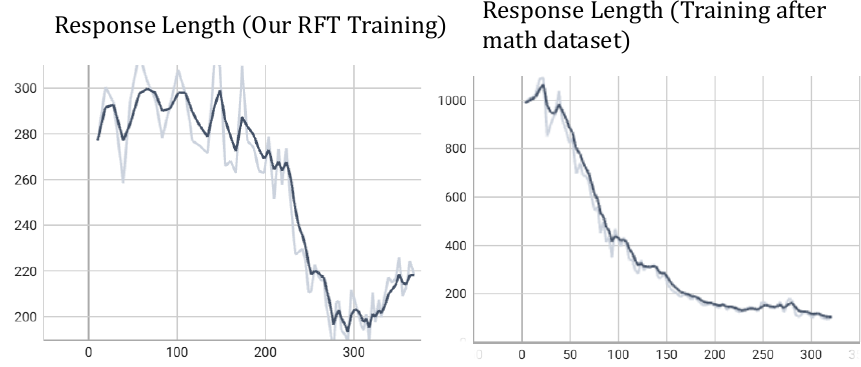}
  \end{minipage}
  \caption{\textbf{Left}:Ablation study on the training stages.\textbf{Right}: Analysis for response length}
  \label{tab:ablation}
\end{table}
\subsection{Ablation Study}

We perform an ablation study to examine the contribution of each training stage in our two-stage pipeline. Specifically, we compare the performance of models trained with only supervised fine-tuning (SFT), only reinforcement fine-tuning (RFT), and the reversed order (RFT before SFT), using the same data and experimental setup. As shown in Table~\ref{tab:ablation}, SFT alone yields substantial improvements on in-domain tasks but fails to generalize to unseen environments. In contrast, applying RFT directly on the base model without prior SFT results in limited gains, likely due to the lack of foundational knowledge. These results confirm the necessity of our two-stage approach, where SFT establishes a strong initialization and RFT enhances generalization through reward-driven optimization.



\subsection{Rethink the reasoning response length:Is Longer always Better?}

In mathematical reasoning tasks, reinforcement fine-tuning often leads to longer outputs without explicit supervision on reasoning length\cite{meng2025mm}. However, we observe that the length of reasoning traces is fundamentally influenced by the nature of the task itself, rather than by a universal tendency toward verbosity. As shown in Figure~\ref{tab:ablation}, our model does not generate longer reasoning outputs after reinforcement fine-tuning; in fact, increased output length does not correlate with higher planning accuracy. To further explore this, we conduct an additional experiment: pretraining the model with mathematical reasoning data to encourage longer responses, then fine-tuning it again on embodied planning data. Although the model initially produces more verbose outputs, the average reasoning length decreases as training progresses, reflecting a domain-specific adaptation toward concise planning.
  
\section{Limitation and Future Work}\label{limitation}

While our work adopts an interactive benchmark for evaluation, the reinforcement fine-tuning process itself does not involve real-time interaction with a simulator. Due to practical constraints in training efficiency and system complexity, we utilize pre-defined truth answers to compute rule-based rewards offline. Incorporating online interaction with the environment to generate learning signals dynamically remains a promising direction for future work, potentially enabling more robust policy refinement through trial-and-error\cite{wang2025ragen}.

In addition, our current focus lies on high-level embodied planning, producing structured action sequences that can guide downstream control modules. Although our method demonstrates strong performance and generalization in simulated benchmarks, it has not yet been deployed on real-world robotic platforms. Extending this framework to physical agents and integrating it with low-level control systems is an important step toward realizing embodied intelligence in practical applications.





\section{Conclusion}

In this paper, we tackle the challenge of enabling vision-language models to perform robust multi-step planning in dynamic embodied environments. To this end, we propose a reinforcement fine-tuning framework that enhances reasoning and decision-making under long-horizon, interactive settings. Our approach combines supervised initialization via knowledge distillation with rule-based reinforcement learning guided by Generalized Reinforced Preference Optimization (GRPO), enabling structured learning from both expert-like demonstrations and task-specific feedback.

We validate our method on Embench, a comprehensive benchmark for interactive embodied planning, demonstrating that our model significantly outperforms both proprietary and open-source baselines of comparable or larger scale. Beyond in-domain performance, our approach shows strong generalization to out-of-distribution tasks and unseen environments. These results highlight the promise of reinforcement-driven reasoning as a scalable and effective direction for advancing embodied intelligence.

\clearpage
\newpage
\bibliographystyle{plainurl}
\bibliography{neurips_2025.bib}

\newpage
\appendix

\section{Appendix Contents}
\begin{itemize}
    \item Section~\ref{appendix_2}: Additional Details of SFT Training Stage
    \item Section~\ref{appendix_3}: Additional Details of RFT Training Stage
    \item Section~\ref{appendix_4}: Additional Details for Evaluation
    \item Section~\ref{appendix_5}: Case study and Visualization
\end{itemize}

\section{Additional Details of SFT Training Stage}\label{appendix_2}

\subsection{Training Dataset Details}

To initialize the model with structured planning priors, we construct a supervised fine-tuning (SFT) dataset via response distillation from a high-capacity proprietary model. Specifically, we query \textit{Gemini-2.0-flash} on a subset of EB-ALFRED tasks from the Embench benchmark and collect its response traces. For each task prompt, we record the full model output, including both intermediate reasoning and executable plan.

To ensure diversity and robustness, we repeat this process across three separate sampling rounds. The final dataset comprises a total of 4,245 training samples. Each sample includes a natural language instruction, egocentric image observations, and the model’s full response in a structured format. An example training instance is shown in figure~\ref{fig:sft_data}

\begin{figure}[!h]
\centering
\begin{minipage}{0.95\linewidth}
\begin{datasetbox}{SFT dataset example}
 {
  "messages": [
   {
    "role": "user",
    "content": {EB-ALFRED prompt}
   },
   {
    "role": "assistant",
    "content": "{"reasoning_and_reflection": "The user wants me to rinse off a ladle and move it to the table. First I need to find the ladle, pick it up, find the sink, put the ladle down in the sink, find the faucet, turn on the faucet to rinse the ladle, turn off the faucet, pick up the ladle, find the table, and put the ladle on the table.",  "visual_state_description": "I can see a garbage can and a fridge in the current view.",  "language_plan": "First, locate a ladle and pick it up. Then, find a sink and put the ladle down in the sink. Next, find a faucet, turn it on to rinse the ladle, and then turn it off. After that, pick up the rinsed ladle, find a dining table, and place the ladle on the table.",  "executable_plan": [   {    "action_id": 64,    "action_name": "find a Ladle"   },   {    "action_id": 109,    "action_name": "pick up the Ladle"   },   {    "action_id": 79,    "action_name": "find a Sink"   },   {    "action_id": 133,    "action_name": "put down the object in hand"   },   {    "action_id": 2,    "action_name": "find a Faucet"   },   {    "action_id": 155,    "action_name": "turn on the Faucet"   },   {    "action_id": 156,    "action_name": "turn off the Faucet"   },   {    "action_id": 64,    "action_name": "find a Ladle"   },   {    "action_id": 109,    "action_name": "pick up the Ladle"   },   {    "action_id": 18,    "action_name": "find a DiningTable"   },   {    "action_id": 133,    "action_name": "put down the object in hand"   }]  }"
    }
  ],
  "images": [
   "example.png"
  ]
 },
\end{datasetbox}
\end{minipage}
\label{fig:sft_data}
\end{figure}

\begin{figure}[!h]
\centering
\begin{minipage}{0.95\linewidth}
\begin{datasetbox}{RFT dataset example}
 {
  {
    "id": "trial_T20190909_062150_965386_remain_0", 
    "question": "{Our_RFT_prompt}",
    "answer": "['Goto handtowelholder', 'Pickup handtowel', 'Goto garbagecan', 'Put handtowel']",
    "message": "[{\"role\": \"system\", \"content\": \"Solve the question. The user asks a question, and you solves it. You first thinks about the reasoning process in the mind and then provides the user with the answer.\"}, {\"role\": \"user\", \"content\": [{\"type\": \"image\", \"image\": \"example.jpg\"}, {\"type\": \"text\", \"text\": \"{Our_RFT_prompt}\"}]}]"
  }
 },
\end{datasetbox}
\end{minipage}
\label{fig:rft_data}
\end{figure}

\subsection{Training Hyperparameters}

We perform full-parameter supervised fine-tuning on the Qwen2.5-VL-7B model using the \texttt{LLaMA-Factory}\cite{zheng2024llamafactory} framework. The training is conducted on 4 NVIDIA A100 40GB GPUs for approximately 8 hours. All hyperparameters are summarized in Table~\ref{tab:sft-hyperparam}.

\begin{table}[!h]
\centering
\small
\setlength{\tabcolsep}{6pt}
\renewcommand{\arraystretch}{1.1}
\begin{tabular}{ll|ll}
\toprule
\textbf{Component} & \textbf{Setting} & \textbf{Component} & \textbf{Setting} \\
\midrule
\multicolumn{4}{c}{\textit{Model Configuration}} \\
image\_max\_pixels & 262144 & freeze\_vision\_tower & true \\
freeze\_language\_model & false & freeze\_multi\_modal\_projector & true \\
deepspeed config & ds\_z3\_config.json &  &  \\
\midrule
\multicolumn{4}{c}{\textit{Dataset Configuration}} \\
dataset & alfred\_sft & template & qwen2\_vl \\
cutoff\_len & 2048 & max\_samples & 1000 \\
overwrite\_cache & true & preprocessing\_workers & 16 \\
dataloader\_workers & 4 &  &  \\
\midrule
\multicolumn{4}{c}{\textit{Training Configuration}} \\
stage & sft & finetuning\_type & full \\
do\_train & true & num\_train\_epochs & 3.0 \\
learning\_rate & 1e-5 & per\_device\_batch\_size & 1 \\
grad\_accum\_steps & 2 & lr\_scheduler & cosine \\
warmup\_ratio & 0.1 & bf16 & true \\
ddp\_timeout & 180000000 &  &  \\
\bottomrule
\end{tabular}
\caption{Detailed hyperparameters used in supervised fine-tuning.}
\label{tab:sft-hyperparam}
\end{table}

\begin{figure}[h]
\centering
\begin{minipage}[t]{0.48\linewidth}
  \vspace{0pt}
  \centering
  \small
  \setlength{\tabcolsep}{5pt}
  \renewcommand{\arraystretch}{1.2}
  \caption{Summary of SFT training results.}
  \vspace{4pt}
  \begin{tabular}{lc}
    \toprule
    \textbf{Metric} & \textbf{Value} \\
    \midrule
    Epochs & 3.0 \\
    Total FLOPs & 3.13e13 \\
    Training Loss & 0.252 \\
    Runtime (s) & 21111.79 \\
    Samples/sec & 0.142 \\
    Steps/sec & 0.018 \\
    \bottomrule
  \end{tabular}
  \label{tab:sft_train_stats}
\end{minipage}
\hfill
\begin{minipage}[t]{0.48\linewidth}
  \vspace{0pt}
  \centering
  \includegraphics[width=\linewidth]{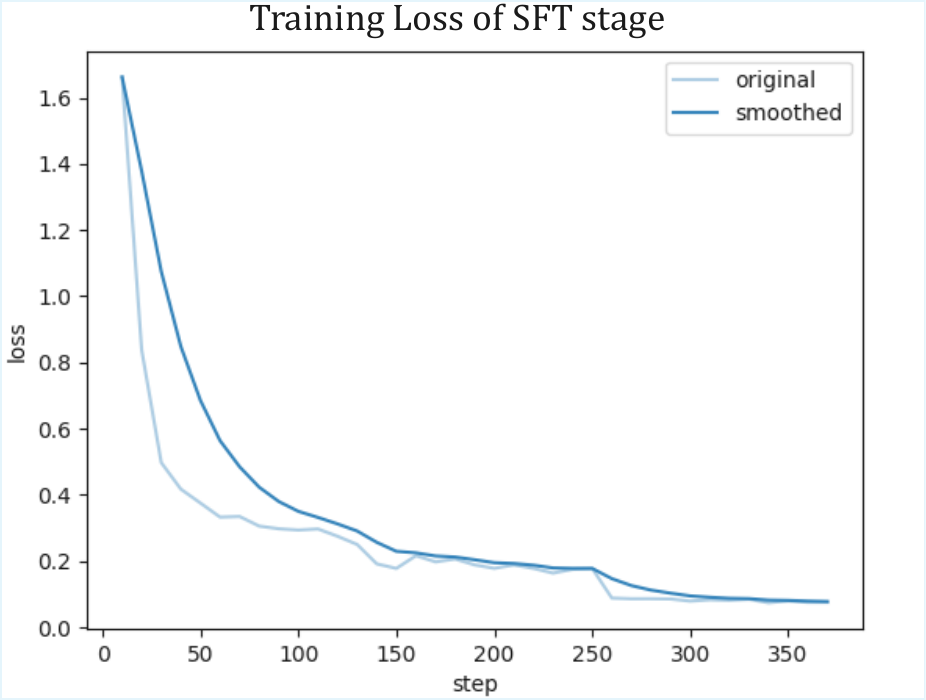} 
  \caption{Training loss curve during SFT stage.}
  \label{fig:sft_loss_curve}
\end{minipage}
\end{figure}

\subsection{Training Results}

We record the final metrics and loss curve from the supervised fine-tuning process, as shown in Figure~\ref{fig:sft_loss_curve}. The table summarizes key training statistics after 3 epochs of full-parameter tuning.

\section{Additional Details of RFT training stage}\label{appendix_3}
\subsection{Training Dataset Details}
We construct our reinforcement fine-tuning (RFT) dataset based on the ALFRED benchmark, following the decomposition and formatting strategy described in Section~\ref{method}. Notably, we do not reuse the SFT-distilled dataset for reinforcement fine-tuning. This decision is motivated by two key considerations: (1) the distilled data may contain suboptimal trajectories, introducing noise into the learning signal; (2) the distilled instruction format is tightly coupled with the benchmark evaluation prompts, whereas our constructed dataset introduces instruction variations that encourage greater policy generalization and better isolate the impact of reinforcement learning.

The resulting dataset contains 43,898 samples, each formatted to include a natural language instruction, a visual observation, and a ground-truth action sequence used for reward computation. We provide a full example of a training sample from the RFT dataset for reference in figure\hyperref[fig:rft_data]{B.2}

\subsection{Training Hyperparameters}
We implement reinforcement fine-tuning using the \texttt{OpenRLHF}\cite{hu2024openrlhf} framework, adopting the Generalized Reinforced Preference Optimization (GRPO) algorithm\cite{shao2024deepseekmath} to optimize policy learning from structured reward feedback. A full list of training hyperparameters is provided in Table~\ref{tab:rft_hyperparams}.

\begin{table}[h]
\centering
\small
\begin{tabularx}{\textwidth}{@{} l X l X @{}}
\toprule
\textbf{Hyperparameter} & \textbf{Value} & \textbf{Hyperparameter} & \textbf{Value} \\
\midrule
ref\_num\_nodes & 1 & vllm\_num\_engines & 8 \\
ref\_num\_gpus\_per\_node & 8 & actor\_num\_gpus\_per\_node & 8 \\
actor\_num\_nodes & 1 & vllm\_tensor\_parallel\_size & 1 \\
vllm\_gpu\_memory\_utilization & 0.65 & vllm\_enable\_sleep & True \\
vllm\_sync\_backend & nccl & temperature & 1.0 \\
max\_epochs & 1 & max\_episodes & 10 \\
prompt\_max\_len & 3000 & max\_samples\_len & 10000 \\
generate\_max\_len & 4096 & advantage\_estimator & group\_norm \\
zero\_stage & 3 & actor\_learning\_rate & 1e-6 \\
init\_kl\_coef & 0.0 & n\_samples\_per\_prompt & 8 \\
micro\_train\_batch\_size & 1 & micro\_rollout\_batch\_size & 2 \\
train\_batch\_size & 128 & rollout\_batch\_size & 128 \\
freeze\_prefix & visual & enable\_accuracy\_filter & True \\
accuracy\_lower\_bound & 0.1 & accuracy\_upper\_bound & 0.9 \\
\bottomrule
\end{tabularx}
\caption{Hyperparameter configuration used during reinforcement fine-tuning.}
\label{tab:rft_hyperparams}
\end{table}

\subsection{Training Log and Result}

\begin{figure}[h]
\centering
\includegraphics[width=0.9\linewidth]{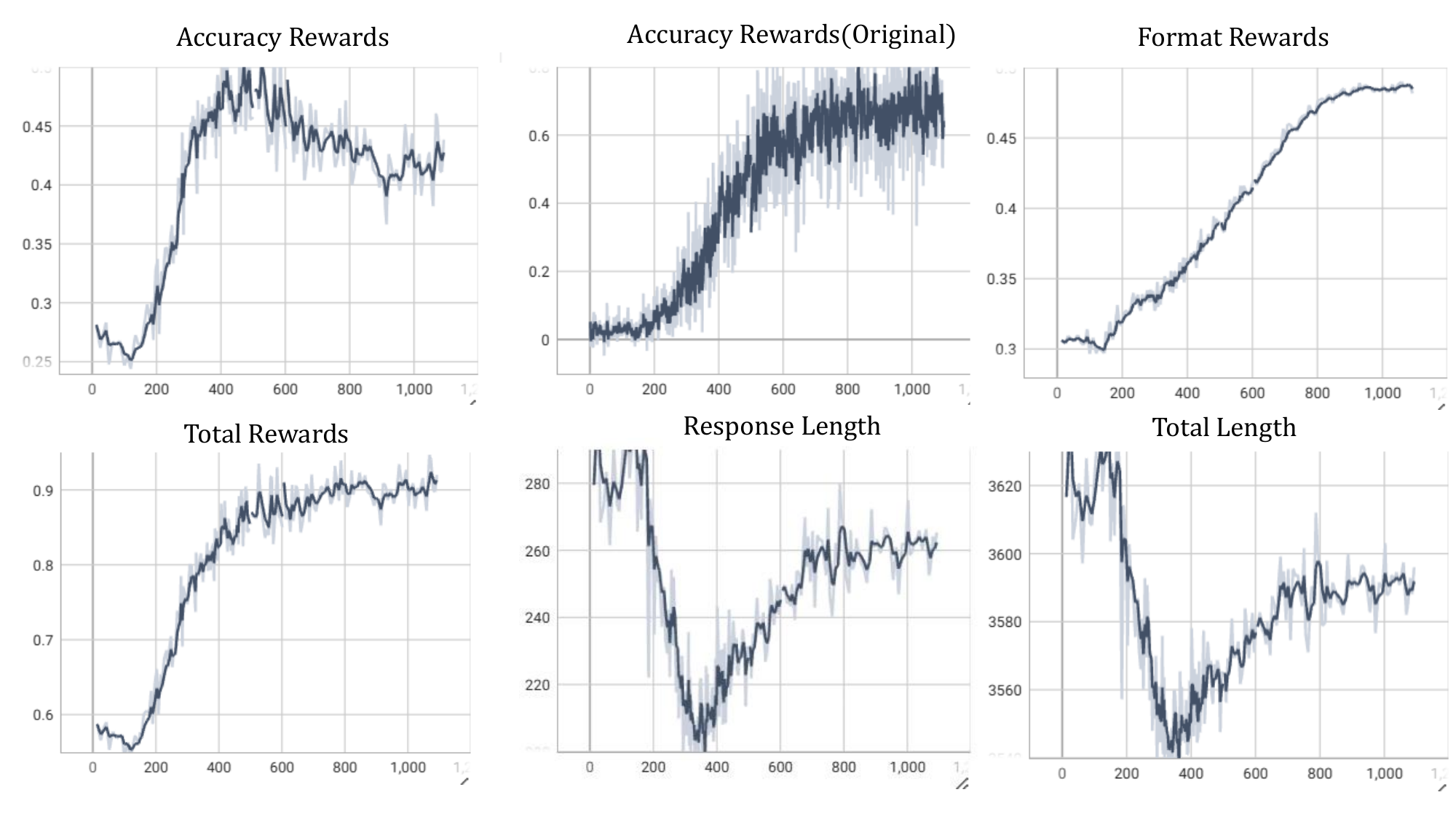}
\caption{Training curve during reinforcement fine-tuning. The figure shows the progression of total reward, filtered and unfiltered accuracy reward, and generation length statistics.}
\label{fig:rft_reward_curve}
\end{figure}

We record the reinforcement fine-tuning process using several key indicators, as visualized in Figure~\ref{fig:rft_reward_curve}.

The \textit{total reward} refers to the combined score of the format reward and the accuracy reward. Due to the use of an online filtering strategy during training, we distinguish between two types of accuracy reward: \textit{accuracy reward (filtered)}, which reflects the reward from selected high-quality samples that pass the filtering criteria, and \textit{accuracy reward (original)}, which represents the average reward across all generated responses prior to filtering.

We also report two types of length statistics: \textit{response length}, which quantifies the number of tokens generated by the model for each output, and \textit{total length}, which denotes the combined token length of the input prompt and generated response. 

\section{Additional Details for Evaluation}\label{appendix_4}

\subsection{Detailed Introduction to EmbodiedBench}

\textbf{EmbodiedBench} is a comprehensive interactive benchmark designed to evaluate vision-language agents in embodied planning scenarios. Unlike static visual question answering settings, EmbodiedBench offers dynamic, simulation-based environments where agents must generate and execute multi-step plans grounded in first-person visual observations and natural language instructions. The benchmark spans four embodied environments and supports over 1,100 diverse tasks with hierarchical action levels, covering both high-level planning and low-level control.

In our work, we focus on two high-level planning environments within EmbodiedBench:

\paragraph{EB-ALFRED.} EB-ALFRED is built upon the ALFRED dataset~\cite{alfred} and implemented on top of the AI2-THOR simulator~\cite{ai2thor}. It supports eight core skill types such as \textit{pick up}, \textit{put down}, \textit{find}, \textit{open/close}, and \textit{turn on/off}. The environment provides egocentric visual inputs and textual feedback (e.g., success/failure messages), enabling agents to adaptively plan and act. Compared to the original ALFRED setup, EB-ALFRED enhances object diversity and simulator robustness. Specifically, it supports multiple object instances of the same type, merges redundant actions (e.g., unified \textit{put down}), and dynamically adjusts the action space size (ranging from 171 to 298). These improvements provide a more realistic and flexible environment for assessing embodied planning capabilities.

\paragraph{EB-Habitat.} EB-Habitat extends the Language Rearrangement benchmark~\cite{habitat}, based on the Habitat 2.0 simulator. It focuses on five high-level skills: \textit{navigation}, \textit{pick}, \textit{place}, \textit{open}, and \textit{close}. Unlike ALFRED, navigation in EB-Habitat is constrained to receptacle-type targets, requiring more sophisticated exploration and scene understanding. The environment includes 282 instruction templates and places more emphasis on spatial reasoning and location-aware planning, making it a complementary testbed for generalization.

\paragraph{Task Subsets.} To enable fine-grained capability analysis, Embench introduces six distinct task subsets:

\begin{itemize}[leftmargin=*]
    \item \textbf{Base:} Evaluates standard task-solving skills under low to medium complexity, testing general planning competence.
    \item \textbf{Common Sense:} Assesses agents’ ability to reason over implicit object references and everyday knowledge.
    \item \textbf{Complex Instruction:} Presents long, noisy or ambiguous contexts to evaluate the agent’s ability to extract user intent.
    \item \textbf{Spatial Awareness:} Requires understanding object relationships in space, such as relative positions or arrangements.
    \item \textbf{Visual Appearance:} Involves identifying objects via attributes like color or shape, testing fine-grained visual recognition.
    \item \textbf{Long Horizon:} Contains tasks demanding long sequences of actions (often exceeding 15 steps), stressing planning depth and temporal consistency.
\end{itemize}

\begin{table*}[htbp]
  \centering
  \small
  \setlength{\tabcolsep}{4pt}
  \renewcommand{\arraystretch}{1.2}
  \caption{Examples of each task type from EB-ALFRED and EB-Habitat.}
  \begin{tabularx}{\textwidth}{l X X}
    \toprule
    \textbf{Task Subset} & \textbf{ALFRED Example} & \textbf{Habitat Example} \\
    \midrule
    Base & Put washed lettuce in the refrigerator. & Move one of the pear items to the indicated sofa. \\
    Common Sense & Place washed leafy green vegetable in a receptacle that can keep it fresh. & Prepare for a game by delivering something to play with to the TV stand. \\
    Complex Instruction & Place the washed lettuce in the refrigerator. This way, it's ready for any delightful recipe ideas you have. & When you find the fridge door open, go ahead and move one bowl to the sofa; otherwise, transport one hammer to the sofa. \\
    Spatial Awareness & Put two spray bottles in the cabinet under the sink against the wall. & Move a spatula from the right counter to the right receptacle of the left counter. \\
    Visual Appearance & Put a knife in a blue container onto the black table in the corner. & Deliver a small red object with green top to the indicated large gray piece of furniture. \\
    Long Horizon & Pick up knife, slice apple, put knife in bowl, heat apple slice in microwave, put apple slice on table. & Move the rubrics cube to the left counter; the towel to the left counter, and the bowl to the brown table. \\
    \bottomrule
  \end{tabularx}
  \label{table:task_type_examples}
\end{table*}

Each subset is designed to probe a specific capability of embodied reasoning, such as commonsense inference, spatial understanding, or long-horizon planning. In our experiments, we evaluate model performance across all six subsets to provide a fine-grained analysis. As shown in Table~\ref{table:task_type_examples}, these categories span a wide range of reasoning challenges. Notably, since our reinforcement fine-tuning dataset only includes \textit{Base} tasks, we observe a significantly larger performance gain in this category, whereas improvements in other subsets are relatively modest. This highlights the need for more diverse training data to support generalizable planning across varied task types.

Overall, Embench provides a rigorous, scalable, and diagnostic framework for benchmarking embodied agents across diverse real-world challenges. In our setup, we use EB-ALFRED for in-domain training and evaluation, while EB-Habitat serves as an out-of-domain testbed to examine generalization performance.

\subsection{Detailed Introduction to Baselines}
To comprehensively evaluate our proposed method, we compare it against a diverse set of baselines, covering both proprietary and open-source models, as well as models specifically optimized for multimodal reasoning and embodied planning. 

(1) \textit{Closed-source models}: we include several leading proprietary vision-language models as strong general-purpose baselines, including Claude-3.5-Sonnet\cite{claude3.5}, Gemini-2.0-flash\cite{gemini2.0}, GPT-4o\cite{gpt-4o}, and GPT-4o-mini\cite{gpt-4o-mini}. 

(2) \textit{Open-source general VLMs}: we evaluate widely adopted open-source VLMs trained for generic multimodal tasks, such as LLaMA-3.2-Vision-11B\cite{llama}, Qwen2.5-VL-7B\cite{qwen2.5} and InternVL2.5-8B\cite{intervl}. 

(3) \textit{Open-source reasoning VLMs}: we further include two representative models that have been explicitly optimized for multimodal reasoning, including MM-Eureka\cite{meng2025mm} and R1-VL\cite{r1-vl}. 

MM-Eureka extends rule-based reinforcement learning to multimodal reasoning, enabling models to improve through reward-driven optimization without supervised fine-tuning. It reproduces key behaviors from language-only RL systems, such as reflection and reward-aligned response growth, achieving strong data efficiency and reasoning performance. 

R1-VL enhances step-by-step reasoning in multimodal LLMs via StepGRPO, a reinforcement learning framework with dense, rule-based rewards for accuracy and logical consistency. It surpasses imitation learning by guiding models to self-correct flawed reasoning, achieving superior results on multiple benchmarks.

We also attempted to evaluate other open-source reasoning models, such as VisualRFT\cite{visualRFT} and Open-R1\cite{openr1}. However, their inference speed was prohibitively slow, resulting in impractically long evaluation time on interactive benchmarks. Additionally, their final planning performance remained poor for embodied planning scenarios.

(4) \textit{Embodied VLMs}: we also include RoboBrain\cite{ji_robobrain_2025} and TAPA\cite{tapa}, two representative open-source large models designed for embodied tasks. 

TAPA is the first model specifically optimized for embodied multi-step planning, but it lacks visual perception capability; thus, we convert visual observations into textual descriptions for evaluation. 

RoboBrain is a state-of-the-art VLM for embodied scenarios that integrates robotic and general multimodal data through a multi-stage training pipeline,leveraging long-horizon video and high-resolution image supervision to enhance manipulation and planning performance.

While there exist other VLMs designed for embodied settings, many of them are unavailable for public use, such as ReasonRFT\cite{reasonrft}, Embodied-R\cite{embodiedR}, and Embodied-Reasoner\cite{EmbodiedReasoner}. Other models, such as EmbodiedGPT\cite{mu_embodiedgpt_2023} and TAPA\cite{tapa}, exhibit poor generalization to new task distributions, achieving near-zero scores on Embench tasks and revealing a lack of transferable planning capabilities.

\subsection{Evaluation Metrics}

We evaluate model performance using \textbf{task success rate}, defined as the percentage of tasks in which the agent successfully completes all required goals. A task is considered successful only if the generated action sequence leads to the environment reaching a final state that satisfies all predefined goal-checking conditions. During evaluation, the vision-language model generates multi-step plans at each interaction step based on the current egocentric observation. If a plan fails—either by producing an invalid action or failing to progress toward the task goal—the agent restarts planning from the latest valid state.

\subsection{Experiment Results using supplementary metrics}

In addition to task success rate, we provide supplementary evaluation results using two additional metrics: \textbf{Progress Rate (PR)} and \textbf{Environment Steps (ES)}.

\textbf{Progress Rate (PR)} quantifies the degree to which the agent completes the task, measured as the proportion of goal conditions satisfied by the final environment state. This metric provides a finer-grained signal than binary success, especially for partially completed tasks.

\textbf{Environment Steps (ES)} refers to the number of actions executed in the environment before task termination. A lower ES generally indicates more efficient planning and fewer redundant or failed actions.

Complete results across these metrics are reported in Appendix Tables~\ref{tab:pr_es_alfred} and~\ref{tab:pr_es_habitat}.
{%
  \setlength{\extrarowheight}{3pt}  
  \setlength{\tabcolsep}{4pt}       

  \begin{table}[htbp]
    \scriptsize                   
    \centering
    \begin{tabular*}{\textwidth}{@{\extracolsep{\fill}}
        >{\centering\arraybackslash}p{2.5cm}  
        *{7}{>{\centering\arraybackslash}p{0.5cm} >{\centering\arraybackslash}p{0.5cm}}
    }
      \toprule
      \textbf{} 
        & \multicolumn{14}{c}{\textbf{EB-ALFRED (Seen)}} \\
      \cmidrule(lr){2-15}
      \textbf{Model} 
      & \multicolumn{2}{c}{\textbf{Avg}} 
      & \multicolumn{2}{c}{\textbf{Base}}
      & \multicolumn{2}{c}{\textbf{Common}}
      & \multicolumn{2}{c}{\textbf{Complex}}
      & \multicolumn{2}{c}{\textbf{Visual}}
      & \multicolumn{2}{c}{\textbf{Spatial}}
      & \multicolumn{2}{c}{\textbf{Long}} \\
      \cmidrule(lr){2-3}  \cmidrule(lr){4-5} \cmidrule(lr){6-7}
      \cmidrule(lr){8-9}  \cmidrule(lr){10-11} \cmidrule(lr){12-13}
      \cmidrule(lr){14-15}
      & \textbf{PR} & \textbf{ES}
      & \textbf{PR} & \textbf{ES}
      & \textbf{PR} & \textbf{ES}
      & \textbf{PR} & \textbf{ES}
      & \textbf{PR} & \textbf{ES}
      & \textbf{PR} & \textbf{ES}
      & \textbf{PR} & \textbf{ES} \\
      \midrule

      \rowcolor{orange!15}
      \multicolumn{15}{c}{\textbf{Closed-Source MLLMs}} \\
      Claude-3.5-Sonnet          
        & 70.11  &  14.9  
        & 72.67  &  12.2   
        & 65.83  &  12.74   
        & 73.33  &  11.48   
        & 65.5   &  14.02   
        & 68.83  &  16.96   
        & 74.5   &  21.98   
        \\
      Gemini-2.0-flash           
        & 57.13 &  16.5
        & 61.83 &  13.96 
        & 60.67 &  14.0 
        & 55.33 &  15.16 
        & 55.33 &  15.26 
        & 46.67 &  17.04 
        & 63.0  &  23.56 
        \\
      GPT-4o                     
        & 61.78 & 16.77 
        & 65.67 &  12.54 
        & 57.17 &  16.1 
        & 74.67 &  13.92 
        & 58.33    &  15.2 
        & 52.33 &  17.58 
        & 62.5    &  25.43
        \\
      GPT-4o-mini                
        & 30.42 &  19.69
        & 36.33 &  17.32 
        & 29.83 &  18.06 
        & 38.0  &  17.74 
        & 27.33 &  18.48 
        & 31.0  &  19.9 
        & 20.0  &  26.62
        \\

      \rowcolor{orange!15}
      \multicolumn{15}{c}{\textbf{Open-Source General MLLMs}} \\
    
      Qwen2.5-VL-7B              
        & 6.86  &  9.4 
        & 5.67  &  8.78 
        & 4.0   &  4.2
        & 5.0   &  5.28
        & 5.33  &  7.16 
        & 0.67  &  8.26 
        & 20.5  &  22.72 
        \\
      InternVL2.5-8B             
        & 5.78  &  7.87
        & 6.17  &  8.2 
        & 0.67  &  4.9 
        & 16.0  &  8.92 
        & 4.0   &  6.78 
        & 6.33  &  7.52 
        & 1.5   &  10.92 
        \\

      \rowcolor{orange!15}
      \multicolumn{15}{c}{\textbf{Open-Source Reasoning MLLMs}} \\
      R1-VL-7B                 
        &  2.78 &  4.01
        &  3.0  &  3.22 
        &  3.0  &  2.06
        &  6.0  &  1.7 
        &  0.67 &  1.62 
        &  0.0  &  2.66
        &  4.0  &  12.78 
        \\
      \makecell[c]{MM-Eureka-Qwen-7B}
        & 6.59  &  8.48
        & 8.67  &  7.64 
        & 5.33  &  5.04
        & 8.67  &  9.72 
        & 3.67  &  6.46 
        & 0.67  &  6.58 
        & 12.5  &  15.42 
        \\

      \rowcolor{orange!15}
      \multicolumn{15}{c}{\textbf{Open-Source Embodied MLLMs}} \\
      RoboBrain                  
        & 1.22  &  6.7 
        & 3.33  &  6.1 
        & 0.67  &  6.3 
        & 0.67  &  3.68 
        & 0.67  &  7.56 
        & 0   &  6.36 
        & 2.0   &  10.22 
        \\
      Tapa                       
        &  0  &  0.03 
        &  0  &  0.06 
        &  0  &  0 
        &  0  &  0
        &  0  &  0.04
        &  0  &  0.08
        &  0  &  0
        \\

      \rowcolor{orange!15}
      \multicolumn{15}{c}{\textbf{Open-Source Embodied + Reasoning MLLMs}} \\
      Ours (Base)                       
        & 6.86  &  9.4 
        & 5.67  &  8.78 
        & 4.0   &  4.2
        & 5.0   &  5.28
        & 5.33  &  7.16 
        & 0.67  &  8.26 
        & 20.5  &  22.72 
        \\
      Ours (SFT only)                       
        & 23.8  & 15.06   
        & 39  &  13.14   
        & 26.6  &  13.04   
        & 27.6  &  12.56   
        & 19.3   &  14.12   
        & 14.3  &  15.16   
        & 16.5   &  22.38   
        \\
      Ours (SFT+RFT)                       
        &  44.25 &  18.53  
        & 61.6  &  15.58   
        & 48.6  &  16.62   
        & 56.7  &  15.94   
        & 38   &  19.06   
        & 42.6  &  17.08   
        & 18   &  26.9   
        \\

      \bottomrule
    \end{tabular*}

    \caption{Progress Rate (PR) and Environment Steps (ES) on EB-ALFRED (Seen)}
    \label{tab:pr_es_alfred}
  \end{table}
}%

{%
  \setlength{\extrarowheight}{3pt}  
  \setlength{\tabcolsep}{4pt}       

  \begin{table}[htbp]
    \scriptsize                   
    \centering
    \begin{tabular*}{\textwidth}{@{\extracolsep{\fill}}
        >{\centering\arraybackslash}p{2.5cm}  
        *{7}{>{\centering\arraybackslash}p{0.5cm} >{\centering\arraybackslash}p{0.5cm}}
    }
      \toprule
      \textbf{} 
        & \multicolumn{14}{c}{\textbf{EB-Habitat (Unseen)}} \\
      \cmidrule(lr){2-15}
      \textbf{Model} 
      & \multicolumn{2}{c}{\textbf{Avg}} 
      & \multicolumn{2}{c}{\textbf{Base}}
      & \multicolumn{2}{c}{\textbf{Common}}
      & \multicolumn{2}{c}{\textbf{Complex}}
      & \multicolumn{2}{c}{\textbf{Visual}}
      & \multicolumn{2}{c}{\textbf{Spatial}}
      & \multicolumn{2}{c}{\textbf{Long}} \\
      \cmidrule(lr){2-3}  \cmidrule(lr){4-5} \cmidrule(lr){6-7}
      \cmidrule(lr){8-9}  \cmidrule(lr){10-11} \cmidrule(lr){12-13}
      \cmidrule(lr){14-15}
      & \textbf{PR} & \textbf{ES}
      & \textbf{PR} & \textbf{ES}
      & \textbf{PR} & \textbf{ES}
      & \textbf{PR} & \textbf{ES}
      & \textbf{PR} & \textbf{ES}
      & \textbf{PR} & \textbf{ES}
      & \textbf{PR} & \textbf{ES} \\
      \midrule

      \rowcolor{green!12}
      \multicolumn{15}{c}{\textbf{Closed-Source MLLMs}} \\
      Claude-3.5-Sonnet          
        & 70.9 &  10.7   
        & 98  &  6.54   
        & 69.5  &  10.46   
        & 75.5  &  10.6   
        & 75.1  &  10.74   
        & 45.2  &  9.44   
        & 62.1  &  16.42   
        \\
      Gemini-2.0-flash           
         & 38.5 &  13.41   
        & 76.5  &  8.56   
        & 31.5  &  12.9   
        & 34  &  15.66   
        & 32.7  &  13.7   
        & 37  &  12   
        & 19.8  &  17.66   
        \\
      GPT-4o                     
         & 60.8 &  14.32   
        & 85.3  &  9.76   
        & 34  &  14.74   
        & 67.5  &  13.34   
        & 64.3  &  13.82   
        & 46.3  &  14.78   
        & 67.2  &  19.5   
        \\
      GPT-4o-mini                
         & 44.2 &  18.8   
        & 73.6  &  10.96   
        & 46  &  18.78   
        & 40.5  &  19.76   
        & 36.8  &  21.76   
        & 47.5  &  18.86   
        & 20.6  &  22.7   
        \\

      \rowcolor{green!12}
      \multicolumn{15}{c}{\textbf{Open-Source General MLLMs}} \\
    
      Qwen2.5-VL-7B              
         & 19.05 &  12.58   
        & 44.5  &  10.64   
        & 6.5  &  14.9   
        & 17  &  11.12   
        & 6.4  &  14.12   
        & 28.8  &  11.74   
        & 11.1  &  12.94   
        \\
      InternVL2.5-8B             
         & 26  & 16.77    
        & 52.9  &  13.1   
        & 13  &  19.1   
        & 22  &  16.48   
        & 21.6  &  18.36   
        & 35.4  &  18.24   
        & 11.1  &  15.32   
        \\

      \rowcolor{green!12}
      \multicolumn{15}{c}{\textbf{Open-Source Reasoning MLLMs}} \\
      R1-VL-7B                 
        & 8.06 &  5.08   
        & 24.6  &  5.9   
        & 0  &  3.78   
        & 4  &  4.38   
        & 6  &  1.8   
        & 11.8  &  7.78   
        & 2  &  6.88   
        \\
      \makecell[c]{MM-Eureka-Qwen-7B}
         & 22.03   &  13.53   
        & 40.5  &  10.24   
        & 20.5  &  15.78   
        & 19  &  11.34   
        & 15.9  &  15.66   
        & 31.3  &  13.74   
        & 5  &  14.4   
        \\

      \rowcolor{green!12}
      \multicolumn{15}{c}{\textbf{Open-Source Embodied MLLMs}} \\
      RoboBrain                  
         & 20.18 & 10.68    
        & 39.1  &  8.08   
        & 9.5  &  9.08   
        & 21  &  11.3   
        & 12.9  &  13.9   
        & 31.1  &  11.48   
        & 7.5  &  10.24   
        \\
      Tapa                       
         & 0 &  0   
        & 0  &  0   
        & 0  &  0   
        & 0  &  0   
        & 0  &  0   
        & 0  &  0   
        & 0  &  0   
        \\

      \rowcolor{green!12}
      \multicolumn{15}{c}{\textbf{Open-Source Embodied + Reasoning MLLMs}} \\
      Ours (Base)               
         & 19.05 &  12.58   
        & 44.5  &  10.64   
        & 6.5  &  14.9   
        & 17  &  11.12   
        & 6.4  &  14.12   
        & 28.8  &  11.74   
        & 11.1  &  12.94   
        \\
      Ours (SFT only)                       
         & 20.05 &   12.40  
        & 38.75  &  10.62   
        & 7  &  12.3   
        & 19.5  &  12.76   
        & 16  &  11.24   
        & 34.6  &  15.26   
        & 4.5  &  12.26   
        \\
      Ours (SFT+RFT)                       
        & 27.18  &  13.31   
        & 58.75  &  8.72   
        & 15  &  14.98   
        & 23  &  13.3   
        & 20  &  13.36   
        & 37  &  13.78   
        & 9.33  & 15.76    
        \\

      \bottomrule
    \end{tabular*}

    \caption{Progress Rate (PR) and Environment Steps (ES) on EB-Habitat (Unseen)}
    \label{tab:pr_es_habitat}
  \end{table}
}%

\subsection{More Ablation Study on RFT Module}
Beyond the primary comparison between supervised fine-tuning (SFT) and reinforcement fine-tuning (RFT), we further conduct ablation studies on two key components within out RFT stage: (1) the \textit{reward allocation curve}, and (2) the \textit{data filtering mechanism}. These modules are designed to enhance the learning signal and training stability.

The reward allocation curve applies a non-linear weighting over step-wise rewards, emphasizing the later steps in a plan trajectory. This design encourages the model to pursue longer-horizon strategies that better reflect complete task execution rather than myopic successes. Meanwhile, the data filtering mechanism discards overly simple or infeasible training samples based on reward thresholds, thereby stabilizing policy learning and preventing overfitting to trivial cases.

Table~\ref{tab:ablation-method} presents the success rate across different task categories in both EB-ALFRED (Seen) and EB-Habitat (Unseen) settings. The removal of either component leads to noticeable performance degradation, underscoring the effectiveness of both modules in our framework.

\begin{table}[ht]
\centering
\small
\rowcolors{2}{gray!10}{white}  
\begin{tabular}{lccccccc}       
\toprule
\textbf{Methods}
  & \textbf{Avg}
  & \textbf{Base}
  & \textbf{Common}
  & \textbf{Complex}
  & \textbf{Visual}
  & \textbf{Spatial}
  & \textbf{Long} \\           
\midrule
\rowcolor{orange!15}
\multicolumn{8}{c}{\textbf{EB-ALFRED (Seen)}} \\  
Ours
  &  \textbf{35.6}    &   54  &   42  &   46  &   28  &   38  &   6 \\
w/o Reward Curve
  & \textbf{35.0} & 50 &  44 &  50 &  34 &  30 &  2 \\
w/o Data Filtering
  & \textbf{25.0} &  42 &  28 &  34 &  26 &  20 &  0 \\
\addlinespace[1ex]
\rowcolor{green!12}
\multicolumn{8}{c}{\textbf{EB-Habitat (Unseen)}} \\  
Ours
  &  \textbf{20}    &  56   &  8   &  18   &  16   &  14   &  8  \\
w/o Reward Curve
  & \textbf{15.33} &  36 &  0 &  12 &  14 &  14 &  6 \\
w/o Data Filtering
  & \textbf{12.6 }&  40 &  4 &  8 &  6 &  14 &  4 \\
\bottomrule
\end{tabular}
\caption{Success Rate (\%) for ablations on the reward allocation curve and data filtering components in the reinforcement fine-tuning stage, evaluated on different task types.}
\label{tab:ablation-method}
\end{table}

\newpage
\section{Case study and Visualization}\label{appendix_5}

\subsection{Case Study}

To better understand how our model performs embodied multi-step planning, we present detailed case studies illustrating its behavior and reasoning process. Specifically, we compare the outputs of our reinforcement-tuned model with the base Qwen2.5-VL model to highlight improvements in planning coherence and action correctness, we also present full multi-step execution trajectories from our model to show how it plans and interacts with the environment to complete specific tasks.

\begin{figure}[htbp]
\centering
    \includegraphics[width=\linewidth]{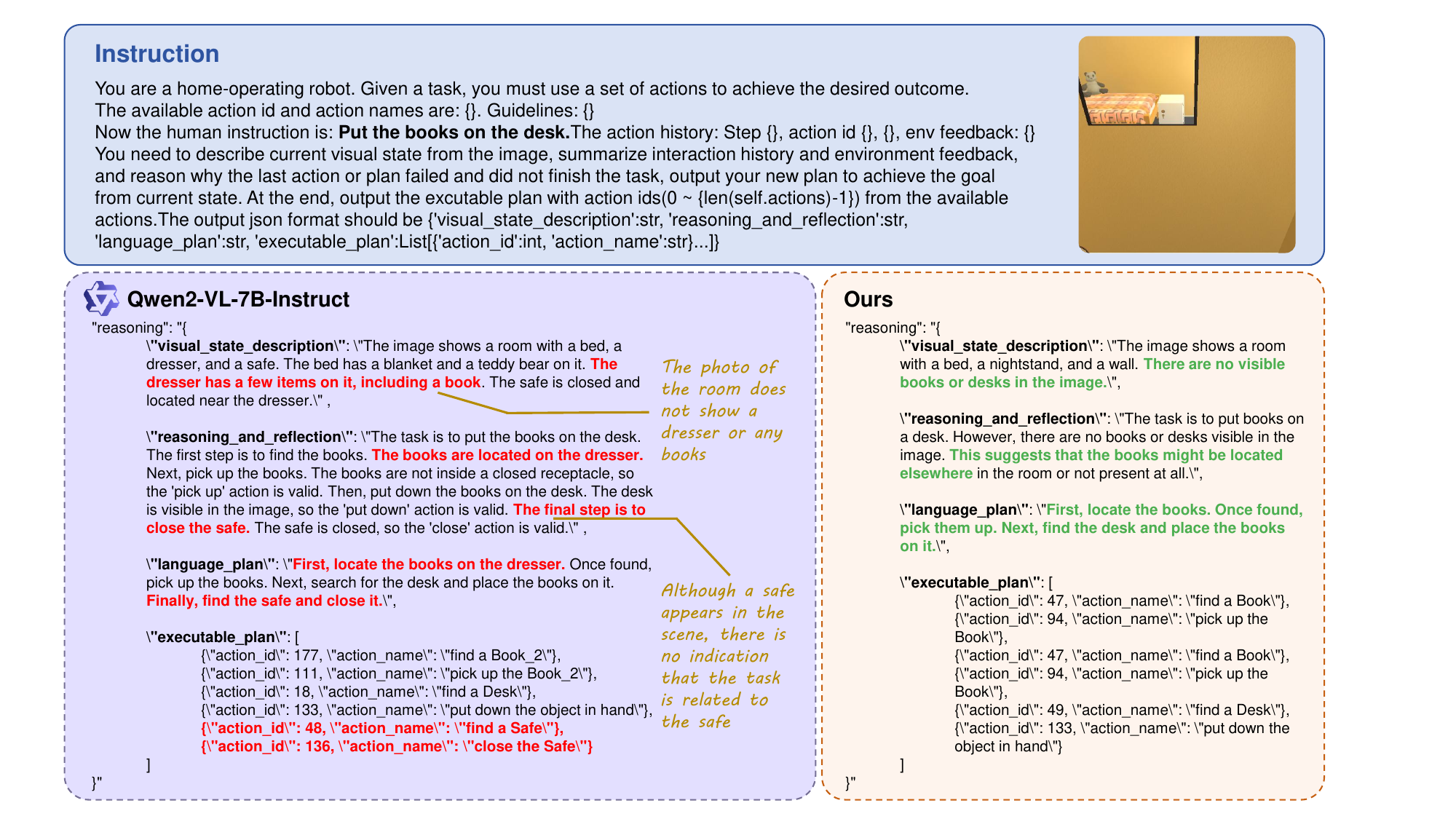}
    \caption{Case comparison (Our model vs Base model) in EB-ALFRED}
    \label{fig:case_study_alf}
\end{figure}

\begin{figure}[htbp]
\centering
    \includegraphics[width=\linewidth]{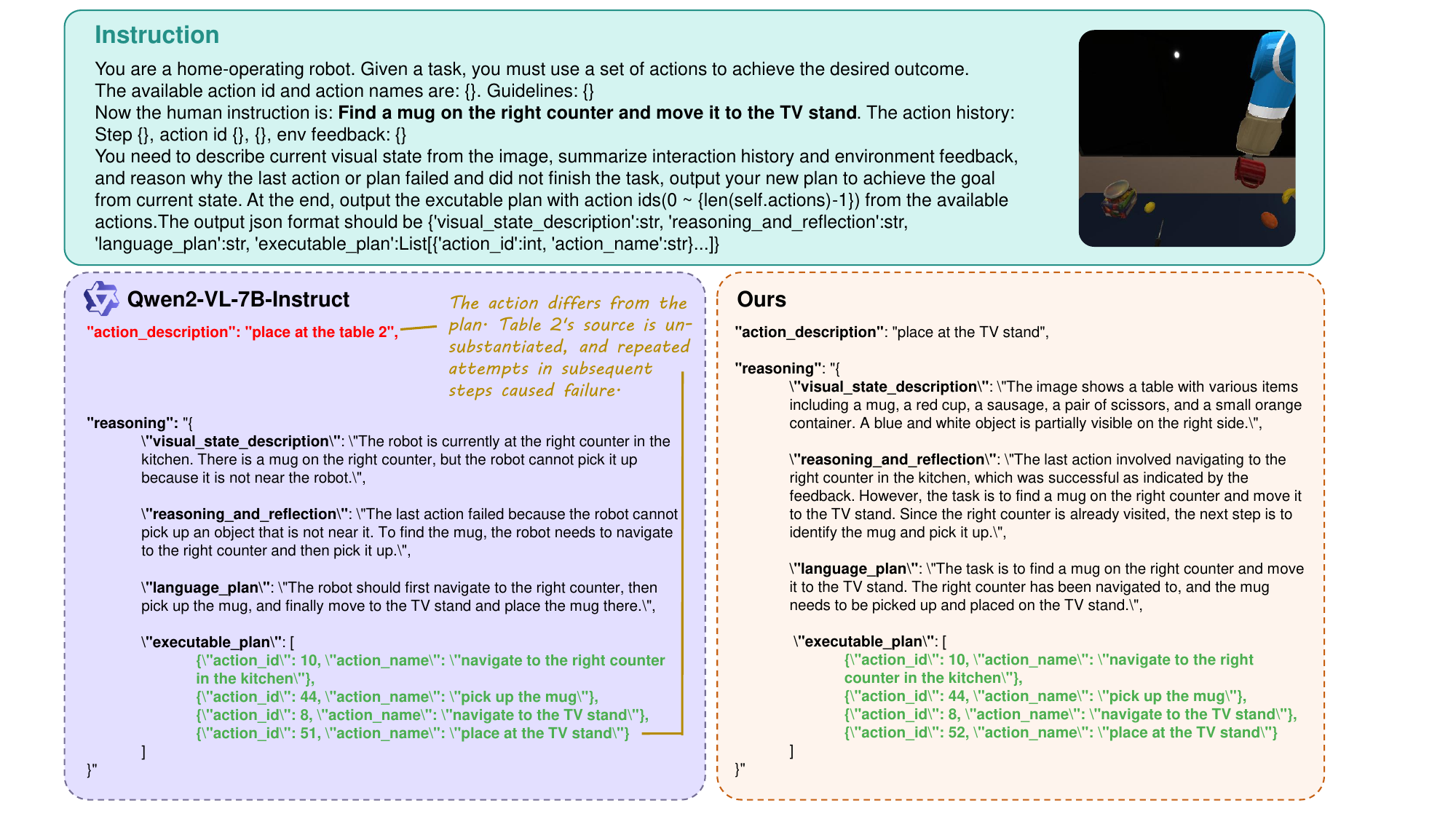}
    \caption{Case comparison (Our model vs Base model) in EB-Habitat}
    \label{fig:case_study_hab}
\end{figure}

Figure~\ref{fig:case_study_alf} and Figure~\ref{fig:case_study_hab} show side-by-side comparisons between the two models in the EB-ALFRED and EB-Habitat environments, respectively. We observe that the base model often produces incomplete or illogical plans, while our model generates more structured and context-aware action sequences, along with interpretable reasoning steps.

\begin{figure}[!t]
\centering
    \includegraphics[width=\linewidth]{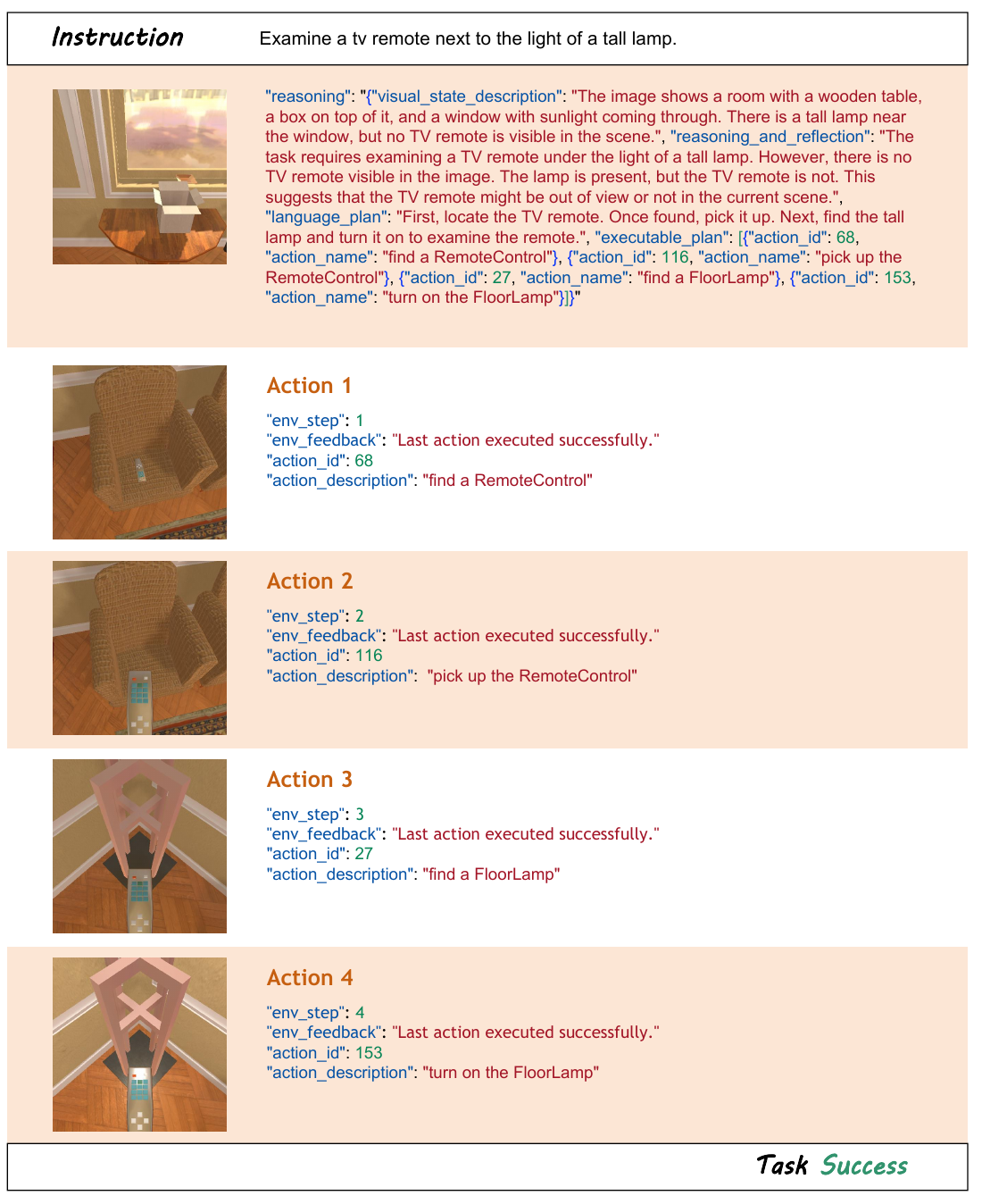}
    \caption{Our model's Full Trajectory execution in EB-ALFRED}
    \label{fig:step_case_study_alf}
\end{figure}

\begin{figure}[!t]
\centering
    \includegraphics[width=\linewidth]{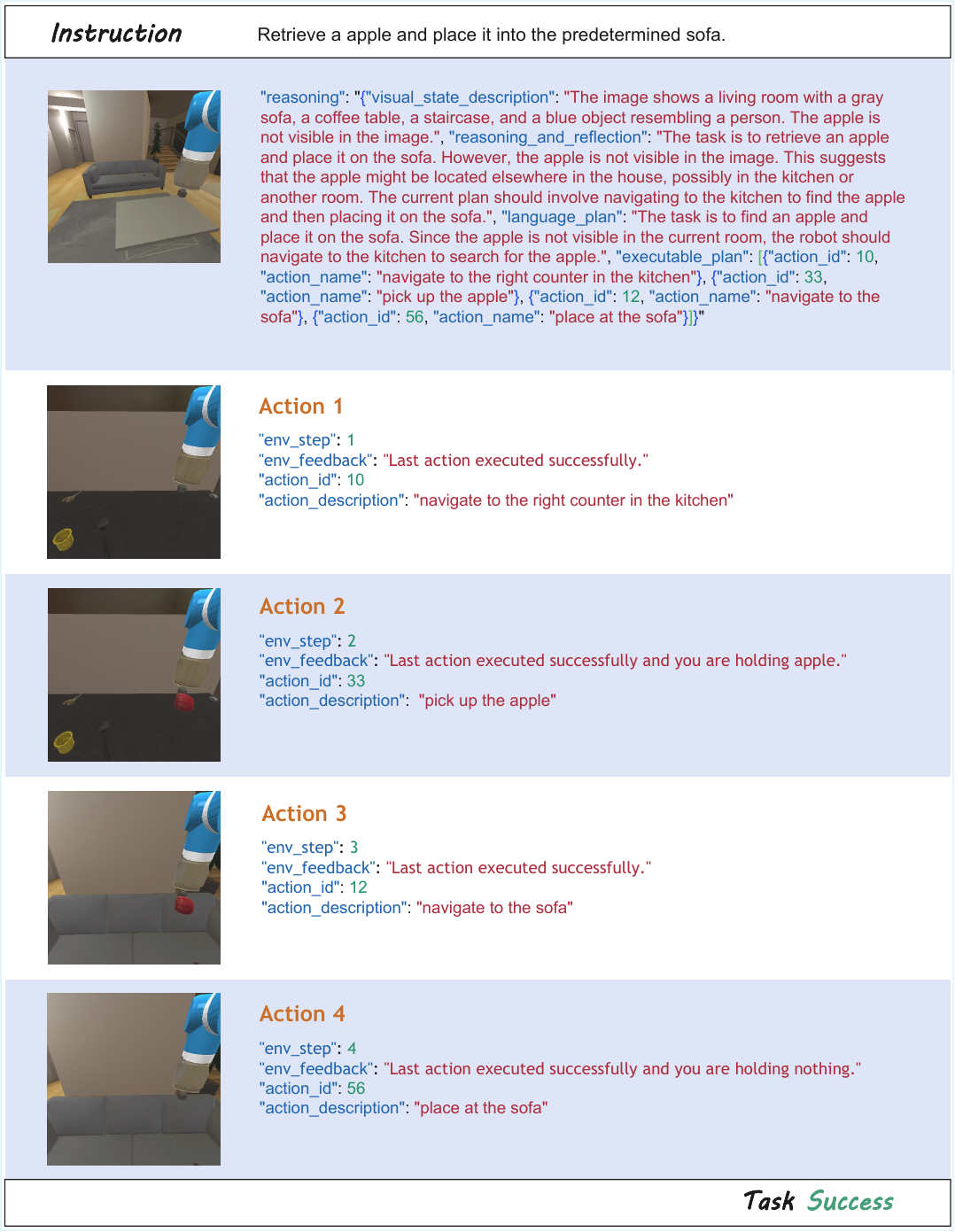}
    \caption{Our model's Full Trajectory execution example 1 in EB-Habitat}
    \label{fig:step_case_study_hab}
\end{figure}

\begin{figure}[!t]
\centering
    \includegraphics[width=\linewidth]{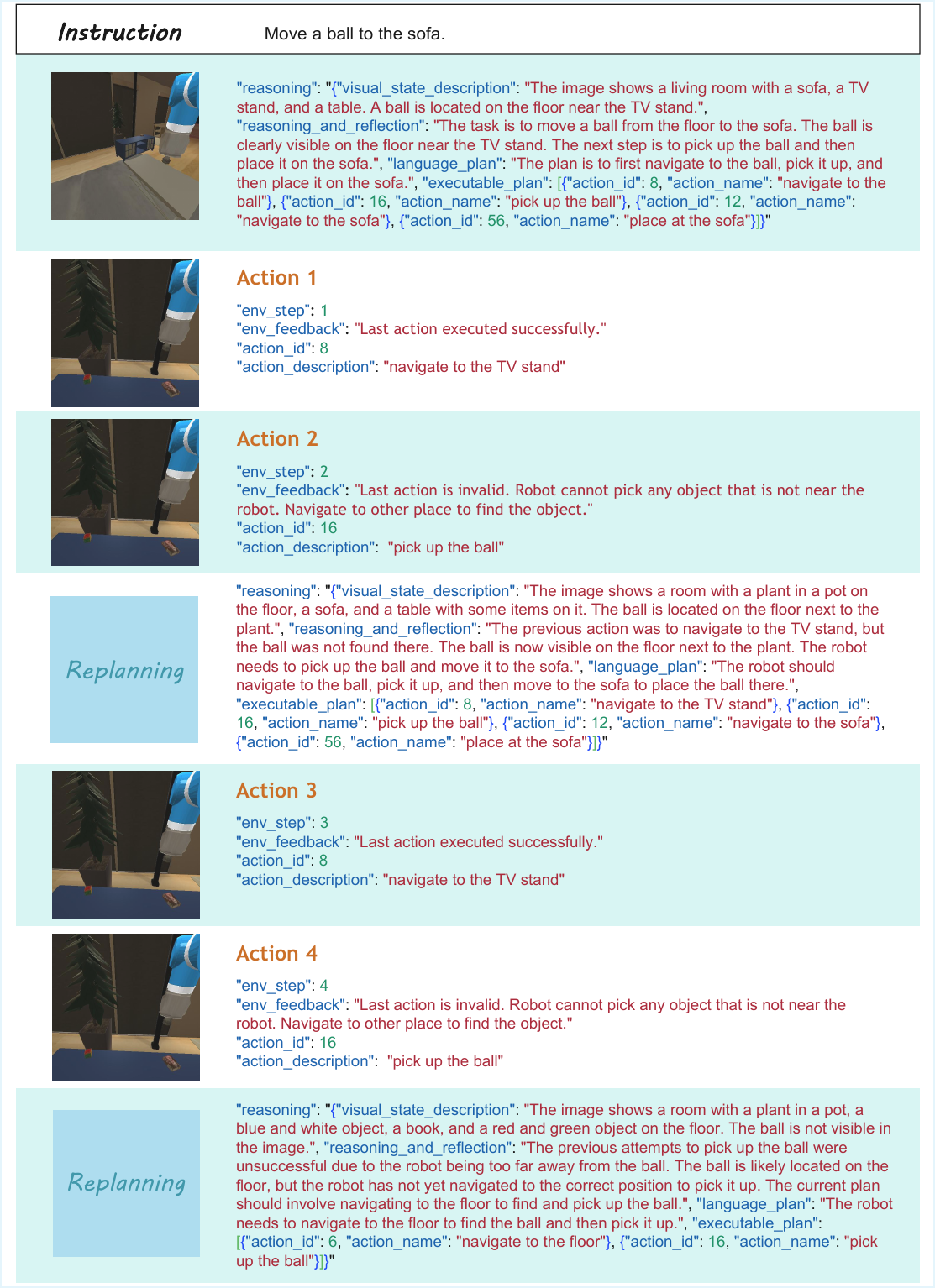}
    \caption{Our model's Full Trajectory execution example 2 (part 1) in EB-Habitat}
    \label{fig:step_case_study_hab2_part1}
\end{figure}

\begin{figure}[!t]
\centering
    \includegraphics[width=\linewidth]{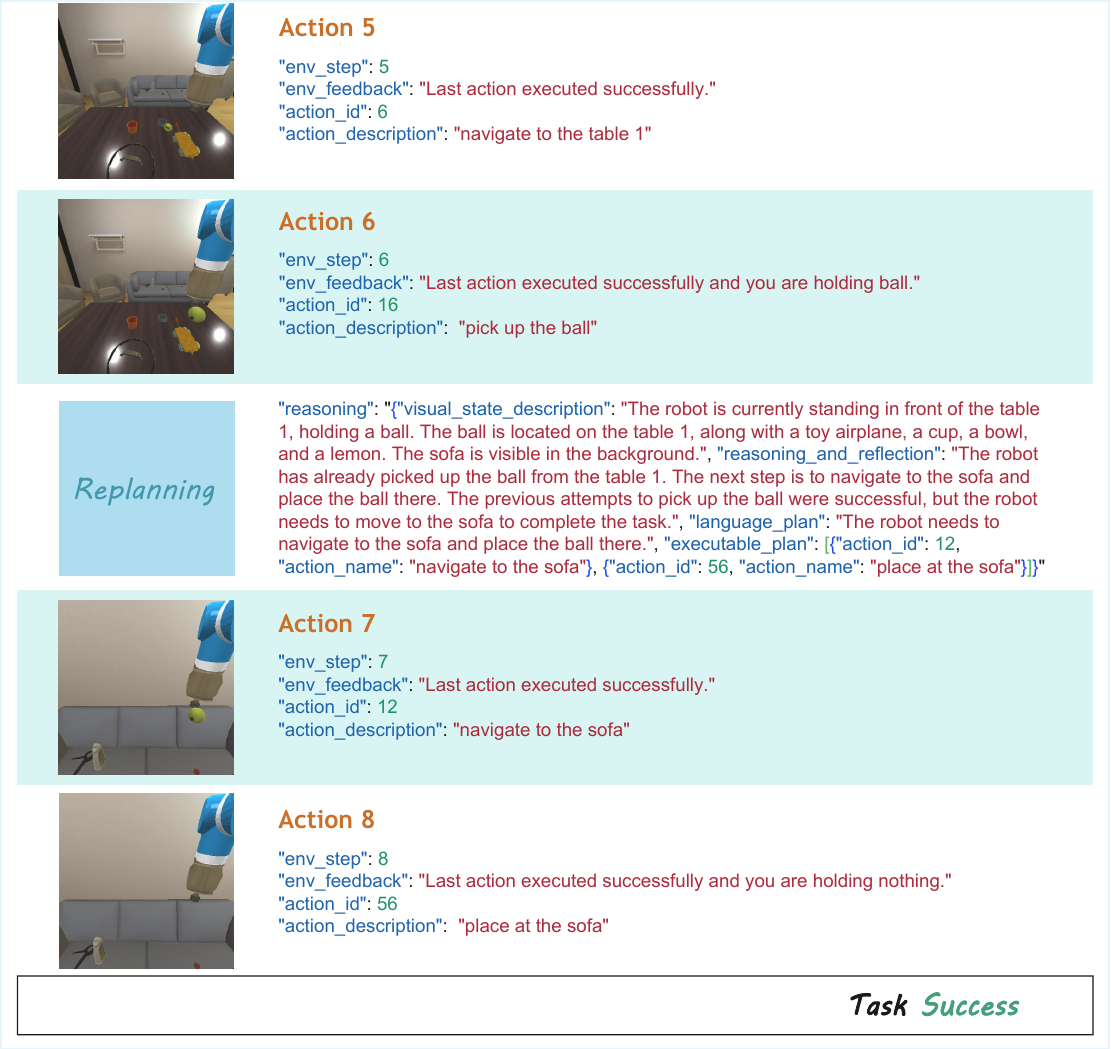}
    \caption{Our model's Full Trajectory execution example 2 (part 2) in EB-Habitat}
    \label{fig:step_case_study_hab2_part2}
\end{figure}

Figure~\ref{fig:step_case_study_alf}, Figure~\ref{fig:step_case_study_hab},Figure~\ref{fig:step_case_study_hab2_part1} and Figure~\ref{fig:step_case_study_hab2_part2} further visualize full planning trajectories executed by our model in representative tasks from EB-ALFRED and EB-Habitat. These examples demonstrate the model’s ability to maintain long-horizon coherence, correctly interpret dynamic observations, and recover from intermediate failures.

\subsection{Prompt}

In this section, we document the full prompt formats used in both evaluation and training stages, including for EB-ALFRED, EB-Habitat, and our reinforcement fine-tuning (RFT) process.

\paragraph{EB-ALFRED Prompt.}  
The EB-ALFRED prompt is used for evaluating models within the EB-ALFRED environment of Embench. Our SFT stage also adopts this prompt format.

\paragraph{EB-Habitat Prompt.}  
This prompt format is used in Embench’s EB-Habitat environment, which differs from EB-ALFRED in simulator, object distribution, and language patterns.

\paragraph{RFT Training Prompt.}  
During reinforcement fine-tuning, we adopt a custom prompt format. While still grounded in the same simulation environment, our RFT prompts include modifications in action representation and instruction phrasing. These differences help introduce broader data diversity and encourage the model to learn a more generalizable planning policy.

\begin{figure}[!h]
\centering
\begin{minipage}{0.95\linewidth}
\begin{promptbox}{EB-ALFRED prompt}
 "<image>## You are a robot operating in a home. Given a task, you must accomplish the task using a defined set of actions to achieve the desired outcome.
 
 ## Action Descriptions and Validity Rules* Find: Parameterized by the name of the receptacle to navigate to. So long as the object is present in the scene, this skill is always valid * Pick up: Parameterized by the name of the object to pick. Only valid if the robot is close to the object, not holding another object, and the object is not inside a closed receptacle.* Put down: Parameterized by the name of the object to put down to a nearby receptacle. Only valid if the robot is holding an object. * Drop: Parameterized by the name of the object to put down. It is different from Put down action, as this does not guarantee the held object will be put into a specified receptacle. * Open: Parameterized by the name of the receptacle to open. Only valid if the receptacle is closed and the robot is close to the receptacle. * Close: Parameterized by the name of the receptacle to close. Only valid if the receptacle is open and the robot is close to the receptacle. * Turn on: Parameterized by the name of the object to turn on. Only valid if the object is turned off and the robot is close to the object. * Turn off: Parameterized by the name of the object to turn off. Only valid if the object is turned on and the robot is close to the object. * Slice: Parameterized by the name of the object to slice. Only valid if the object is sliceable and the robot is close to the object.
 
 ## The available action id (0 ~ 207) and action names are: {ALFRED ACTION LIST}

 ## Task Execution Example:{IN-CONTEXT TASK EXAMPLE}

 ## Guidelines 1. **Output Plan**: Avoid generating empty plan. Each plan should include no more than 20 actions. 2. **Visibility**: Always locate a visible object by the 'find' action before interacting with it. 3. **Action Guidelines**: Make sure match the action name and its corresponding action id in the output.\n Avoid performing actions that do not meet the defined validity criteria. For instance, if you want to put object in a receptacle, use 'put down' rather than 'drop' actions.  4. **Prevent Repeating Action Sequences**: Do not repeatedly execute the same action or sequence of actions.  Try to modify the action sequence because previous actions do not lead to success. 5. **Multiple Instances**: There may be multiple instances of the same object, distinguished by an index following their names, e.g., Cabinet_2, Cabinet_3. You can explore these instances if you do not find the desired object in the current receptacle. 6. **Reflection on History and Feedback**: Use interaction history and feedback from the environment to refine and improve your current plan. If the last action is invalid, reflect on the reason, such as not adhering to action rules or missing preliminary actions, and adjust your plan accordingly.
 
 ## Now the human instruction is: Rinse off a ladle and move it to the table. You are supposed to output in json. You need to describe current visual state from the image, output your reasoning steps and plan. At the end, output the action id (0 ~ 207) from the available actions to excute."
\end{promptbox}
\end{minipage}
\label{fig:eb-alf-prompt}
\end{figure}

\begin{figure}[!h]
\centering
\begin{minipage}{0.95\linewidth}
\begin{promptbox}{EB-Habitat prompt}
  <image>##You are a robot operating in a home. Given a task, you must accomplish the task using a defined set of actions to achieve the desired outcome.
  
  ## Action Descriptions and Validity Rules: * Navigation: Parameterized by the name of the receptacle to navigate to. So long as the receptacle is present in the scene, this skill is always valid. * Pick: Parameterized by the name of the object to pick. Only valid if the robot is close to the object, not holding another object, and the object is not inside a closed receptacle. * Place: Parameterized by the name of the receptacle to place the object on. Only valid if the robot is close to the receptacle and is holding an object. * Open: Parameterized by the name of the receptacle to open. Only valid if the receptacle is closed and the robot is close to the receptacle. * Close: Parameterized by the name of the receptacle to close. Only valid if the receptacle is open and the robot is close to the receptacle.
  
  ## The available action id (0 ~ 69) and action names are:{HABITAT ACTION LIST}

  ## Task Execution Example:{IN-CONTEXT TASK EXAMPLE}

  ## Guidelines 1. **Output Plan**: Avoid generating empty plan. Each plan should include no more than 20 actions. 2. **Visibility**: If an object is not currently visible, use the \"Navigation\" action to locate it or its receptacle before attempting other operations. 3. **Action Validity**: Make sure match the action name and its corresponding action id in the output. Avoid performing actions that do not meet the defined validity criteria. 4. **Prevent Repeating Action Sequences**: Do not repeatedly execute the same action or sequence of actions. Try to modify the action sequence because previous actions do not lead to success. 5. **Multiple Instances**: There may be multiple instances of the same object, distinguished by an index following their names, e.g., cabinet 2, cabinet 3. You can explore these instances if you do not find the desired object in the current receptacle. 6. **Reflection on History and Feedback**: Use interaction history and feedback from the environment to refine and enhance your current strategies and actions. If the last action is invalid, reflect on the reason, such as not adhering to action rules or missing preliminary actions, and adjust your plan accordingly.
  
  ## Now the human instruction is: Move one of the pear items to the indicated sofa. You are supposed to output in json. You need to describe current visual state from the image, output your reasoning steps and plan. At the end, output the action id (0 ~ 69) from the available actions to excute."

  
\end{promptbox}
\end{minipage}
\label{fig:eb-hab-prompt}
\end{figure}

\begin{figure}[!h]
\centering
\begin{minipage}{0.95\linewidth}
\begin{promptbox}{Our RFT prompt}
 You are a robot operating in a home. Given a task, you must accomplish the task using a defined set of actions to achieve the desired outcome.
 
 ## Action Descriptions and Validity Rules * GotoLocation: Parameterized by the name of the target location or receptacle to navigate to. Always valid so long as the target exists in the scene. * PickupObject: Parameterized by the name of the object to pick up. Valid only if the robot is close to the object, is not holding anything, and the object is accessible. * PutObject: Parameterized by the name of the receptacle or surface where the held object will be placed. Valid only if the robot is holding an object. * ToggleObject: Parameterized by the name of the object whose state can be toggled (e.g., lamp, faucet). Valid only if the robot is close to the object. * CoolObject: Parameterized by the name of the object to cool. Requires the robot to be holding the object and near a cooling appliance such as a fridge. * SliceObject: Parameterized by the name of the object to slice. Requires that the object is slice-able and the robot holds an appropriate cutting tool. * CleanObject: Parameterized by the name of the object to clean. Requires the robot to be near a water source and the object supports cleaning. * HeatObject: Parameterized by the name of the object to heat. Requires the robot to be holding the object and near a heating appliance such as a microwave or stove.
 
 ## The available action id (0 ~ 224) and action names are:{OUR RFT ACTION LIST}

 ## Guidelines 1. **Output Plan**: Avoid generating empty plan. Each plan should include no more than 20 actions. 2. **Visibility**: Always locate a visible object by the 'goto' action before interacting with it. 3. **Action Guidelines**: Make sure the action name and its corresponding action id match in the output. Avoid performing actions that do not meet the defined validity criteria. 4. **Prevent Repeating Action Sequences**: Do not repeatedly execute the same action or sequence of actions. 5. **Multiple Instances**: There may be multiple instances of the same object, distinguished by an index following their names, e.g., Cabinet_2. 6. **Reflection on History and Feedback**: Use interaction history and feedback from the environment to refine and improve your current plan.
 
 ## Expected JSON output format```json {\"reasoning_and_reflection\": \"<string>\",  \"visual_state_description\": \"<string>\",  \"language_plan\": \"<string>\", \"executable_plan\": [ {\"action_id\": <int>, \"action_name\": \"<string>\"} ]}```
 
## Now the human instruction is: put a towel into a garbage can The history actions are: [{HISTORY LIST}] \nConsidering the above interaction history and the current image state, to achieve the human instruction.\nYou are supposed to output in json. You need to describe current visual state from the image, output your reasoning steps and plan. You shuold think carefully and output the comprehensive thought process in 'reasoning_and_reflection' part. At the end, output the action id (0 ~ 224) from the available actions to execute."
\end{promptbox}
\end{minipage}
\label{fig:our-rft prompt}
\end{figure}

\begin{figure}[!h]
\centering
\begin{minipage}{0.95\linewidth}
\begin{listbox}{Part of EB-ALFRED Action list}
 action id 1: find a Potato, action id 2: find a Faucet, action id 3: find a Ottoman, action id 4: find a CoffeeMachine, action id 5: find a Candle, action id 6: find a CD, action id 7: find a Pan, action id 8: find a Watch, action id 9: find a HandTowel, action id 10: find a SprayBottle, action id 11: find a BaseballBat, action id 12: find a CellPhone, action id 13: find a Kettle, action id 14: find a Mug, action id 15: find a StoveBurner, action id 16: find a Bowl, action id 17: find a Toilet, action id 18: find a DiningTable, action id 19: find a Spoon, action id 20: find a TissueBox, action id 21: find a Shelf, action id 22: find a Apple, action id 23: find a TennisRacket, action id 24: find a SoapBar, action id 25: find a Cloth, action id 26: find a Plunger, action id 27: find a FloorLamp, action id 28: find a ToiletPaperHanger, action id 29: find a CoffeeTable, action id 30: find a Spatula, action id 31: find a Plate, action id 32: find a Bed, action id 33: find a Glassbottle, action id 34: find a Knife, action id 35: find a Tomato, action id 36: find a ButterKnife, action id 37: find a Dresser, action id 38: find a Microwave, action id 39: find a CounterTop, action id 40: find a GarbageCan, action id 41: find a WateringCan, action id 42: find a Vase, action id 43: find a ArmChair, action id 44: find a Safe, action id 45: find a KeyChain, action id 46: find a Pot, action id 47: find a Pen, action id 48: find a Cabinet, action id 49: find a Desk, action id 50: find a Newspaper, action id 51: find a Drawer, action id 52: find a Sofa, action id 53: find a Bread, action id 54: find a Book, action id 55: find a Lettuce, action id 56: find a CreditCard, action id 57: find a AlarmClock, action id 58: find a ToiletPaper, action id 59: find a SideTable, action id 60: find a Fork, action id 61: find a Box, action id 62: find a Egg, action id 63: find a DeskLamp, action id 64: find a Ladle, action id 65: find a WineBottle, action id 66: find a Pencil, action id 67: find a Laptop, action id 68: find a RemoteControl, action id 69: find a BasketBall, action id 70: find a DishSponge, action id 71: find a Cup, action id 72: find a SaltShaker, action id 73: find a PepperShaker, action id 74: find a Pillow, action id 75: find a Bathtub, action id 76: find a SoapBottle, action id 77: find a Statue, action id 78: find a Fridge, action id 79: find a Sink, action id 80: pick up the KeyChain, action id 81: pick up the Potato, action id 82: pick up the Pot, action id 83: pick up the Pen, action id 84: pick up the Candle, action id 85: pick up the CD, action id 86: pick up the Pan, action id 87: pick up the Watch, action id 88: pick up the Newspaper, action id 89: pick up the HandTowel, action id 90: pick up the SprayBottle, action id 91: pick up the BaseballBat, action id 92: pick up the Bread, action id 93: pick up the CellPhone, action id 94: pick up the Book, action id 95: pick up the Lettuce, action id 96: pick up the CreditCard, action id 97: pick up the Mug, action id 98: pick up the AlarmClock, action id 99: pick up the Kettle, action id 100: pick up the ToiletPaper
\end{listbox}
\end{minipage}
\label{fig:eb-alf-prompt-list}
\end{figure}

\begin{figure}[!h]
\centering
\begin{minipage}{0.95\linewidth}
\begin{listbox}{EB-Habitat Action list}
 action id 0: navigate to the cabinet 7, action id 1: navigate to the cabinet 6, action id 2: navigate to the cabinet 5, action id 3: navigate to the cabinet 4, action id 4: navigate to the refrigerator push point, action id 5: navigate to the chair 1, action id 6: navigate to the table 1, action id 7: navigate to the table 2, action id 8: navigate to the TV stand, action id 9: navigate to the sink in the kitchen, action id 10: navigate to the right counter in the kitchen, action id 11: navigate to the left counter in the kitchen, action id 12: navigate to the sofa, action id 13: navigate to the refrigerator, action id 14: navigate to the left drawer of the kitchen counter, action id 15: navigate to the right drawer of the kitchen counter, action id 16: pick up the ball, action id 17: pick up the clamp, action id 18: pick up the hammer, action id 19: pick up the screwdriver, action id 20: pick up the padlock, action id 21: pick up the scissors, action id 22: pick up the block, action id 23: pick up the drill, action id 24: pick up the spatula, action id 25: pick up the knife, action id 26: pick up the spoon, action id 27: pick up the plate, action id 28: pick up the sponge, action id 29: pick up the cleanser, action id 30: pick up the plum, action id 31: pick up the pear, action id 32: pick up the peach, action id 33: pick up the apple, action id 34: pick up the lemon, action id 35: pick up the can, action id 36: pick up the box, action id 37: pick up the banana, action id 38: pick up the strawberry, action id 39: pick up the lego, action id 40: pick up the rubriks cube, action id 41: pick up the book, action id 42: pick up the bowl, action id 43: pick up the cup, action id 44: pick up the mug, action id 45: pick up the orange, action id 46: pick up the lid, action id 47: pick up the toy airplane, action id 48: pick up the wrench, action id 49: place at the chair 1, action id 50: place at the table 1, action id 51: place at the table 2, action id 52: place at the TV stand, action id 53: place at the sink in the kitchen, action id 54: place at the right counter in the kitchen, action id 55: place at the left counter in the kitchen, action id 56: place at the sofa, action id 57: place at the refrigerator, action id 58: place at the left drawer of the kitchen counter, action id 59: place at the right drawer of the kitchen counter, action id 60: open the refrigerator, action id 61: close the refrigerator, action id 62: open the cabinet 7, action id 63: open the cabinet 6, action id 64: open the cabinet 5, action id 65: open the cabinet 4, action id 66: close the cabinet 7, action id 67: close the cabinet 6, action id 68: close the cabinet 5, action id 69: close the cabinet 4
\end{listbox}
\end{minipage}
\label{fig:eb-hab-prompt-list}
\end{figure}

\begin{figure}[!h]
\centering
\begin{minipage}{0.95\linewidth}
\begin{listbox}{Part of Our RFT Action list}
 action id 1: goto apple, action id 2: goto armchair, action id 3: goto baseballbat, action id 4: goto basketball, action id 5: goto bathtubbasin, action id 6: goto bed, action id 7: goto bowl, action id 8: goto box, action id 9: goto bread, action id 10: goto butterknife, action id 11: goto cabinet, action id 12: goto candle, action id 13: goto cart, action id 14: goto cellphone, action id 15: goto cloth, action id 16: goto coffeemachine, action id 17: goto coffeetable, action id 18: goto countertop, action id 19: goto creditcard, action id 20: goto cup, action id 21: goto desk, action id 22: goto desklamp, action id 23: goto diningtable, action id 24: goto dishsponge, action id 25: goto drawer, action id 26: goto dresser, action id 27: goto egg, action id 28: goto floorlamp, action id 29: goto fork, action id 30: goto fridge, action id 31: goto garbagecan, action id 32: goto handtowelholder, action id 33: goto keychain, action id 34: goto knife, action id 35: goto laptop, action id 36: goto lettuce, action id 37: goto microwave, action id 38: goto mug, action id 39: goto newspaper, action id 40: goto ottoman, action id 41: goto pan, action id 42: goto pen, action id 43: goto pencil, action id 44: goto plate, action id 45: goto plunger, action id 46: goto pot, action id 47: goto potato, action id 48: goto remotecontrol, action id 49: goto safe, action id 50: goto shelf, action id 51: goto sidetable, action id 52: goto sinkbasin, action id 53: goto soapbar, action id 54: goto soapbottle, action id 55: goto sofa, action id 56: goto spatula, action id 57: goto spoon, action id 58: goto statue, action id 59: goto stoveburner, action id 60: goto tennisracket, action id 61: goto tissuebox, action id 62: goto toilet, action id 63: goto toiletpaper, action id 64: goto toiletpaperhanger, action id 65: goto tomato, action id 66: goto vase, action id 67: goto watch, action id 68: goto wateringcan, action id 69: pickup alarmclock, action id 70: pickup apple, action id 71: pickup baseballbat, action id 72: pickup basketball, action id 73: pickup book, action id 74: pickup bowl, action id 75: pickup box, action id 76: pickup bread, action id 77: pickup butterknife, action id 78: pickup candle, action id 79: pickup cd, action id 80: pickup cellphone, action id 81: pickup cloth, action id 82: pickup creditcard, action id 83: pickup cup, action id 84: pickup dishsponge, action id 85: pickup egg, action id 86: pickup fork, action id 87: pickup glassbottle, action id 88: pickup handtowel, action id 89: pickup kettle, action id 90: pickup keychain, action id 91: pickup knife, action id 92: pickup ladle, action id 93: pickup laptop, action id 94: pickup lettuce, action id 95: pickup mug, action id 96: pickup newspaper, action id 97: pickup pan, action id 98: pickup pen, action id 99: pickup pencil, action id 100: pickup peppershaker, 
\end{listbox}
\end{minipage}
\label{fig:our-rft-prompt-list}
\end{figure}

\end{document}